%% file: iclr2026_conference.tex
\documentclass{article} 
\usepackage{iclr2026_conference,times}
\usepackage{xspace}
\usepackage{booktabs}
\usepackage{graphicx}
\usepackage{multirow}
\usepackage{tabularx,adjustbox}
\usepackage{subcaption}
\usepackage{wrapfig}
\usepackage[most]{tcolorbox}
\usepackage{pifont}
\usepackage[T1]{fontenc}
\usepackage{lmodern}
\input{math_commands.tex}

\usepackage{hyperref}
\usepackage{url}
\usepackage[table]{xcolor}
\usepackage{enumitem}

\newcommand{\ours}{MemoryAgentBench\xspace}

\definecolor{cream}{RGB}{255, 251, 234}
\definecolor{goldenstar}{RGB}{255, 204, 0}
\definecolor{midnightgreen}{rgb}{0.0, 0.29, 0.33}
\definecolor{deepgreen}{HTML}{0aa344}
\definecolor{deeppurple}{HTML}{7030a0}
\definecolor{deepblue}{HTML}{171d91}
\definecolor{brown}{HTML}{843c0c}
\definecolor{shadered}{HTML}{ffe5e5}
\definecolor{shadegreen}{HTML}{e5f7ed}
\definecolor{msftBlack}{RGB}{0,0,0}
\definecolor{lightred}{RGB}{255,163,163}
\definecolor{deepred}{RGB}{146,0,0}
\definecolor{headergray}{RGB}{240,240,240}
\definecolor{groupblue}{RGB}{220,235,250}
\definecolor{groupgreen}{RGB}{220,250,220}
\definecolor{grouppink}{RGB}{250,230,240}
\definecolor{groupyellow}{RGB}{255,250,205}
\definecolor{groupgray}{RGB}{245,245,245}
\definecolor{avgcolor}{RGB}{255, 245, 220} 
\definecolor{overallcolor}{RGB}{255, 235, 205} 

\title{Evaluating Memory in LLM Agents via Incremental Multi-Turn Interactions}

\author{
Yuanzhe Hu\thanks{These authors contributed equally to this work.}\thanks{Joint corresponding authors.}*†, Yu Wang*†, Julian McAuley \\
University of California, San Diego \\
\texttt{\{yuh127, yuw164, jmcauley\}@ucsd.edu}
\\
\\
\raisebox{-0.25em}{
  \includegraphics[width=0.033\linewidth]{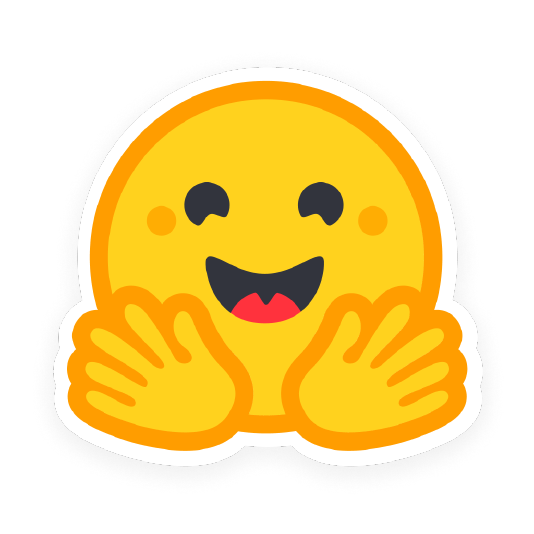}%
} \href{https://huggingface.co/datasets/ai-hyz/MemoryAgentBench}{Datasets}\quad
\raisebox{-0.1em}{%
  \includegraphics[width=0.025\linewidth]{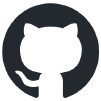}%
} \href{https://github.com/HUST-AI-HYZ/MemoryAgentBench}{Source Code}
\vspace{-0.5cm}
}

\iclrfinalcopy 

\begin{document}

\maketitle

\begin{abstract}
Recent benchmarks for Large Language Model (LLM) agents primarily focus on evaluating reasoning, planning, and execution capabilities, while another critical component—memory, encompassing how agents memorize, update, and retrieve long-term information—is under-evaluated due to the lack of benchmarks. We term agents with memory mechanisms as \textbf{memory agents}. 
In this paper, based on classic theories from memory science and cognitive science, we identify four core competencies essential for memory agents: accurate retrieval, test-time learning, long-range understanding, and selective forgetting. Existing benchmarks either rely on limited context lengths or are tailored for static, long-context settings like book-based QA, which do not reflect the interactive, multi-turn nature of memory agents that incrementally accumulate information. Moreover, no existing benchmarks
cover all four competencies. 
We introduce \textbf{\ours}, a new benchmark specifically designed for memory agents. Our benchmark transforms existing long-context datasets and incorporates newly constructed datasets into a multi-turn format, effectively simulating the incremental information processing characteristic of memory agents. By carefully selecting and curating datasets, our benchmark provides comprehensive coverage of the four core memory competencies outlined above, thereby offering a systematic and challenging testbed for assessing memory quality. 
We evaluate a diverse set of memory agents, ranging from simple context-based and retrieval-augmented generation (RAG) systems to advanced agents with external memory modules and tool integration. Empirical results reveal that current methods fall short of mastering all four competencies, underscoring the need for further research into comprehensive memory mechanisms for LLM agents.
\end{abstract}

\input{sections/introduction}
\input{sections/related_work}

\input{sections/method}

\input{sections/experiments}
\input{sections/conclusion}

\newpage
\input{sections/statement}

\bibliography{iclr2026_conference}
\bibliographystyle{iclr2026_conference}

\newpage

\appendix
\input{sections/appendix}

\end{document}

%% file: math_commands.tex
\usepackage{amsmath,amsfonts,bm}

\def\eqref#1{equation~\ref{#1}}

\def\1{\bm{1}}

\DeclareMathAlphabet{\mathsfit}{\encodingdefault}{\sfdefault}{m}{sl}
\SetMathAlphabet{\mathsfit}{bold}{\encodingdefault}{\sfdefault}{bx}{n}



%% file: sections/introduction.tex
\section{Introduction}
\label{sec:introduction}
Large Language Model (LLM) agents have rapidly transitioned from proof-of-concept chatbots to end-to-end systems that can write software~\citep{openhands2024}, control browsers~\citep{browser_use2024}, and reason over multi-modal inputs. Frameworks such as \textsc{Manus}, \textsc{OWL}~\citep{owl2025}, \textsc{OpenHands}~\citep{openhands2024}, and \textsc{Codex} routinely solve complex, tool-rich tasks and achieve state-of-the-art results on agentic benchmarks like GAIA~\citep{gaia} and SWE-Bench~\citep{swebench}. Yet these evaluations focus almost exclusively on \emph{reasoning}  (planning, tool using, code synthesis) and leave the equally important question of \emph{memorization} (abstraction, storing, updating, retrieving) largely under-explored. Recent memory-centric architectures—ranging from parametric memory systems like MemoryLLM~\citep{memoryllm}, SELF-PARAM~\citep{self-param}, and M+~\citep{m+} to commercial token-level memory solutions such as \textsc{MemGPT}~\citep{MemGPT,sleep-time-compute}, \textsc{Mem0}~\citep{Mem0}, \textsc{Cognee}~\citep{cognee}, \textsc{Zep}~\citep{Zep} and \textsc{MIRIX}~\citep{mirix}—employ diverse strategies for storing and retrieving past information. Despite growing interest, their real-world effectiveness remains largely anecdotal, and there is currently no unified benchmark for systematically evaluating the quality of memory in agents. In this paper, we refer to agents equipped with memory mechanisms as \textbf{Memory Agents}, where memory can take various forms, including parameters, vectors, textual histories, or external databases. In this paper, we primarily focus on memory agents that utilize textual histories and external databases, as these approaches are most commonly deployed in real-world applications. In contrast, memory encoded in model parameters~\citep{memoryllm,m+,yin2024explicit} remains largely within academic research and is typically less capable than proprietary memory systems equipped on closed-sourced API models.

\begin{figure}
    \centering
    \includegraphics[width=0.9\linewidth]{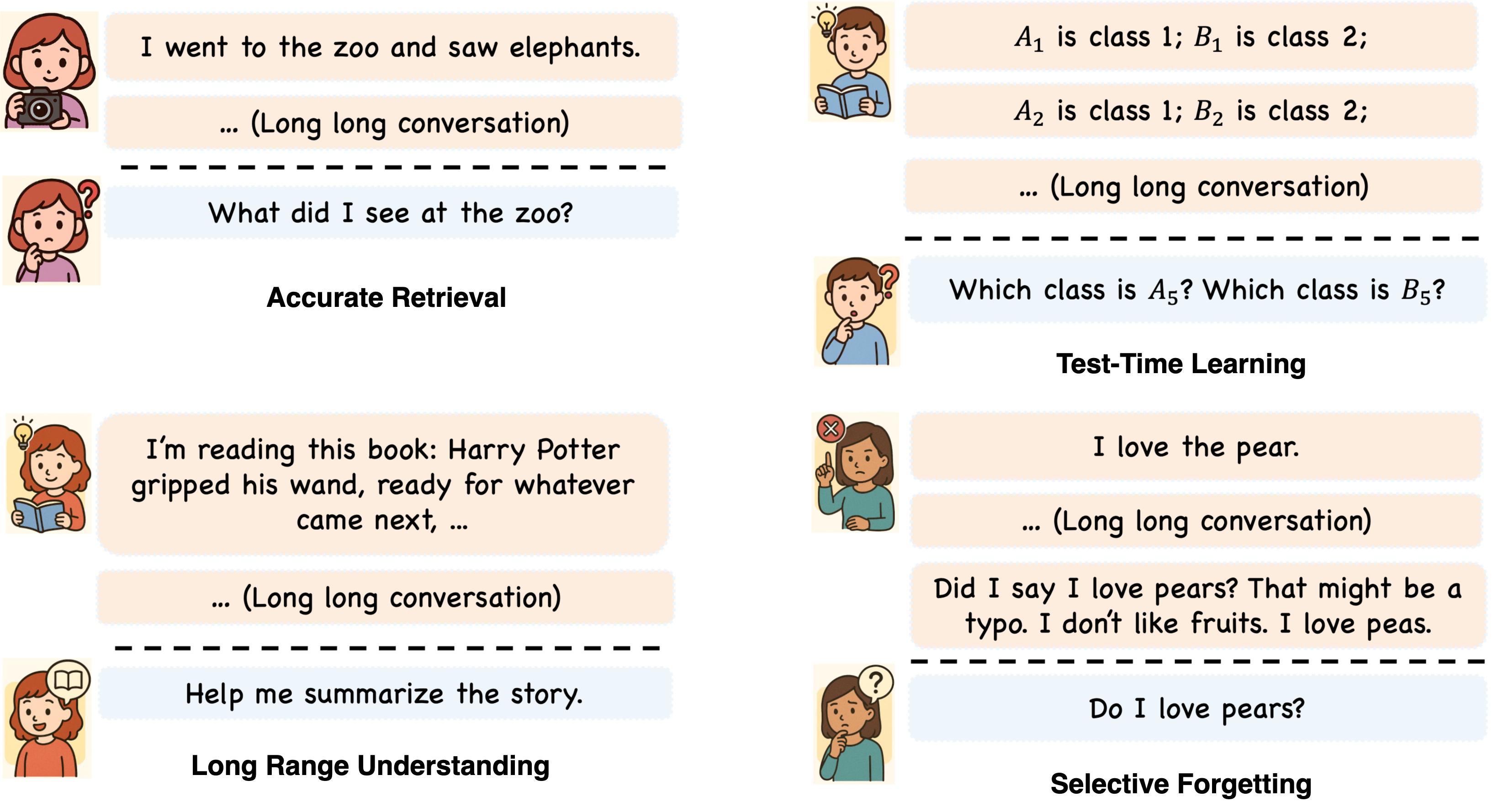}
    \caption{Four complementary competencies that memory agents should have.}
    \label{fig:intro}
    \vspace{-10pt}
\end{figure}

Based on some classic theories in memory and cognitive science~\citep{James1890PrinciplesV1,McClelland1995CLS,AndersonNeely1996Interference,Wimber2015AdaptiveForgetting}, we identify four complementary competencies (Examples shown in Figure~\ref{fig:intro}) to evaluate memory agents:
(1) \textbf{Accurate Retrieval (AR)}: The ability to extract the correct snippet in response to a query. This can involve one-hop or multi-hop retrieval, as long as the relevant information can be accessed with a single query.
(2) \textbf{Test-Time Learning (TTL)}: The capacity to incorporate new behaviors or acquire new skills during deployment, without additional training.
(3) \textbf{Long-Range Understanding (LRU)}: The ability to integrate information distributed across extended contexts ($\geq$ 100k tokens) and answer questions requiring a global understanding of the entire sequence.
(4) \textbf{Selective Forgetting (SF)}: The skill to revise, overwrite, or remove previously stored information when faced with contradictory evidence, aligning with goals in model editing and knowledge unlearning tasks~\citep{memit,Large_Scale_Knowledge_Washing}. For these four competencies, we provide more detailed definitions in Appendix ~\ref{sec:details_of_datasets}.

Previous datasets developed to evaluate memory in language models have notable limitations. Early benchmarks such as \textsc{LOCOMO}~\citep{locomo} ($\sim$9k tokens), LooGLE\citep{LooGLE} ($\sim$24k tokens), and LongBench\citep{longbench} ($\sim$ 20k tokens) feature relatively short contexts that no longer challenge current models. More recent datasets like NovelQA\citep{novelqa} ($\sim$200k tokens), NOCHA\citep{nocha} ($\sim$127k tokens), Loong\citep{loong} ($\sim$100k tokens), and $\infty$-Bench\citep{infinitebench} ($\sim$150k tokens) extend the context length to evaluate global reasoning and retrieval capabilities. However, these datasets were primarily designed for evaluating long-context language models rather than memory agents. 
The reason that long-context benchmarks cannot be directly used to evaluate memory agents is as follows. There is a fundamental distinction between memory and long context: memory serves as a compressed and distilled representation of past information. Rather than storing all historical content verbatim, memory selectively extracts salient details, removes irrelevant information, and often incorporates new inferences derived from prior experiences. Consequently, \textbf{memory agents are designed to process context incrementally}—absorbing input piece by piece, abstracting and consolidating information over time, generating new inferences, and learning novel rules from accumulated history. For this reason, datasets that provide the entire context in a single block are not directly applicable to evaluating memory agents. A more recent effort, \textsc{LongMemEval}~\citep{longmemeval}, seeks to address this limitation by using synthetic long-form conversations, which can be injected into memory gradually, session by session. Nonetheless, its evaluation framework remains constrained by limited topical diversity and less realistic interaction patterns, reducing its applicability to real-world memory agent scenarios.

To address these limitations, we introduce a unified benchmark, \textbf{\ours}, specifically designed to evaluate a broad spectrum of memory mechanisms in agent systems. We also provide a framework for memory agent evaluation. In this framework, agents are presented with sequences of textual inputs that simulate multi-turn interactions with users. We reconstructed existing datasets originally developed for long-context LLM evaluation by segmenting and reconstructing inputs into multiple dialogue chunks and feeding them incrementally to the agent in a time order. However, since these datasets do not fully capture all four targeted memory competencies, we also introduce two new datasets: \textbf{EventQA} and \textbf{FactConsolidation}, designed to evaluate accurate retrieval and selective forgetting, respectively.
Our benchmark includes evaluations of state-of-the-art commercial memory agents (such as MIRIX and MemGPT), long-context agents that treat the full input as memory, and RAG agents that extend their memory through retrieval methods. We examine how techniques developed for long-context models and RAG transfer to the memory agent setting. By providing a consistent evaluation protocol across diverse agent architectures and datasets, \ours delivers comprehensive insights into agent performance across the four core memory competencies. In Table~\ref{tab:benchmark-comparison}, we compare the \ours with previous representative benchmarks across multiple dimensions.


Our contributions are summarized as follows:
\begin{itemize}[leftmargin=*]
\item \textbf{Datasets:} We reconstruct existing datasets and create two new datasets to construct a comprehensive benchmark, covering four distinct memory competencies. 
\item \textbf{Framework:} We provide a unified evaluation framework, and open-source the codebase and datasets to encourage reproducibility and further research. 
\item \textbf{Empirical Study:} We implement various simple agents with diverse memory mechanisms, adopt commercial agents, and evaluate these agents on our proposed benchmark. Our results demonstrate that existing memory agents, although effective in some tasks, still face significant challenges in certain aspects.
\end{itemize}

%% file: sections/related_work.tex
\section{Related Work}

\subsection{Benchmarks on Long-Context and Memory} 

In this section, we review prior work on evaluation benchmarks, categorizing them into three domains: long-context understanding, retrieval-augmented generation, and memory agents. 

\vspace{-10pt}

\paragraph{Benchmarks for Long-Context LLMs.} Early benchmarks designed for long-context evaluation include LongBench\citep{longbench} and LooGLE\citep{LooGLE}, with average input lengths of approximately 20k and 24k tokens, respectively. More recent benchmarks—such as $\infty$-Bench~\citep{infinitebench}, HELMET\citep{HELMET}, RULER\citep{RULER}, NOCHA\citep{nocha}, NoLiMa~\citep{nolima} and LongBench V2\citep{longbenchv2}—extend context lengths to over 100k tokens. While these benchmarks effectively assess the model's ability to process extensive information in a single pass, they are primarily intended for static long-context reading comprehension and do not reflect the incremental, multi-turn nature of memory agents.

\paragraph{Benchmarks for Retrieval-Augmented Generation.}
Beyond pure long-context evaluation, a line of benchmarks targets retrieval-augmented generation (RAG) for knowledge-intensive tasks such as open-domain QA, fact checking, and document ranking over fixed corpora, e.g., KILT~\citep{kilt} and BEIR~\citep{beir}. 
More recent work explicitly evaluates end-to-end RAG systems under long-context or application-specific scenarios, including LaRA~\citep{lara}, LONG$^2$RAG~\citep{LONG2RAG}, FRAMES~\citep{FRAMES}, and CRUD-RAG~\citep{Crud-rag}. Large-scale benchmarks such as RAGBench~\citep{ragbench}, RAGTruth~\citep{ragtruth}, FreshLLM~\citep{freshllms}, and $T^2$-RAGBench~\citep{t2ragbench} further extend the evaluation space to industrial manuals, hallucination detection, time-sensitive web QA, and text-and-table financial reports, respectively.
However, existing RAG benchmarks typically assume a static or slowly changing knowledge base and short-lived interactions, emphasizing retrieval accuracy and grounding over the continual updating and selective forgetting of information that is central to memory agents.

\paragraph{Benchmarks for Memory Agents}

More recently, benchmarks such as LOCOMO~\citep{locomo}, LongMemEval~\citep{longmemeval}, RealTalk~\citep{Realtalk} and StoryBench~\citep{StoryBench} have been proposed specifically for evaluating memory agents. While promising, LOCOMO still features relatively short conversations ($\sim$9k), and LongMemEval uses synthetic conversations with limited topical diversity, making the dialogues less realistic and potentially less representative of real-world memory use cases. Meanwhile, the evaluation scope of the above benchmarks is not sufficient to comprehensively assess the four core competencies—Accurate Retrieval, Test-Time Learning, Long-Range Understanding, and Selective Forgetting—that are essential for robust memory agents.

\begin{table}[!t]
\centering
\caption{A comparison between \ours and existing long-term memory QA benchmarks. \#Q denote the total number of questions. Context depth is defined as the number of tokens in the history. *Not reported in the paper, based on our approximation. The context depth of StoryBench is not reported in paper. We compare these datasets in terms of their ability to \textbf{comprehensively and effectively evaluate} the each capability dimension that we propose. We also compare prior work in terms of their evaluation coverage of memory agents—specifically, whether they provide comprehensive assessments across different categories of memory methods: Long-Context Agents (LCA), RAG Agents, and Agentic Memory (AM). }

\resizebox{0.96\linewidth}{!}{
\begin{tabular}{c|c|c|c|c|c|c|c|c|c}
\hline
\multirow{2}{*}{\textbf{Benchmark}}   & \multirow{2}{*}{\textbf{\#Q}} & \multirow{2}{*}{\textbf{Context Depth}} & \multicolumn{4}{c|}{\textbf{Core Memory Competencies}} & \multicolumn{3}{c}{\textbf{Evaluation Coverage}} \\ 
             &  &  & \multicolumn{1}{c}{\textbf{AR}} & \multicolumn{1}{c}{\textbf{TTL}} & \multicolumn{1}{c}{\textbf{LRU}} & \multicolumn{1}{c|}{\textbf{SF}} & \multicolumn{1}{c}{\textbf{LCA}} & \multicolumn{1}{c}{\textbf{RAG}} & \multicolumn{1}{c}{\textbf{AM}}  \\
\hline
MemoryBank ~\citep{MemoryBank}   & 194 &  5k             & \checkmark & \checkmark  & \ding{55} & \ding{55} & \checkmark & \ding{55} & \checkmark  \\
LoCoMo ~\citep{locomo}     & 7512 &  10k  & \checkmark & \ding{55} & \ding{55} & \ding{55} & \checkmark & \checkmark & \ding{55}   \\
PerLTQA~\citep{perltqa}    & 8593 & 1M* & \checkmark & \ding{55} & \ding{55} & \ding{55} & \checkmark & \checkmark & \ding{55}   \\
RealTalk ~\citep{Realtalk}   & 728 &  375k* & \checkmark & \ding{55} & \checkmark  & \ding{55} & \checkmark & \ding{55} & \ding{55}   \\
LongMemEval ~\citep{longmemeval}        & 500 &  115k,  1.5M   & \checkmark & \ding{55} & \ding{55} & \ding{55} & \checkmark & \checkmark & \ding{55}   \\
StoryBench ~\citep{StoryBench}    & 86 &  -  & \checkmark & \ding{55} & \checkmark & \ding{55} & \checkmark  & \ding{55} & \ding{55}   \\
\ours        & 2071 &  103k-1.44M   & \checkmark & \checkmark & \checkmark & \checkmark & \checkmark & \checkmark & \checkmark   \\

\hline
\end{tabular}
}

\label{tab:benchmark-comparison}
\end{table}

\subsection{Agents with Memory Mechanisms}
\label{sub:agents_with_memory_mechanisms}
Memory mechanisms are attracting more and more attention lately~\citep{lscs}.
Recent advancements in LLMs have demonstrated the capability to process extended context lengths, ranging from 100K to over 1 million tokens. For instance, models such as GPT-4o~\citep{openai2025gpt4o} and Claude 3.7~\citep{anthropic2025claude37} can handle inputs of approximately 100K to 200K tokens, while models like Gemini 2.0 Pro~\citep{deepmind2025gemini} and the GPT-4.1 series extend this capacity beyond 1 million tokens. These strong long-context capabilities enable a simple yet effective form of memory: storing information directly within the context window. However, this approach is inherently constrained by a hard limit—once the context window is exceeded, earlier information must be discarded.

In parallel, RAG continues to serve as a dominant paradigm for managing excessive context. By retrieving relevant information from earlier context and feeding it to the LLM, RAG allows systems to overcome context length limitations. For example, OpenAI’s recent memory functionality\footnote{\url{https://openai.com/index/memory-and-new-controls-for-chatgpt/}} combines explicit user preference tracking with retrieval-based methods that reference prior interactions. RAG methods can be broadly classified into three categories:
\textbf{Simple RAG}: These methods rely on string-matching techniques such as TF-IDF, BM25~\citep{bm25}, and BMX~\citep{BMX}, which are entirely non-neural and operate on string-level similarity.
\textbf{Embedding-based RAG}: This class leverages neural encoders, primarily transformers, to map text into dense vector representations~\citep{wu2022pcl}. Early methods like DPR~\citep{karpukhin2020dense} and Contriever~\citep{contriever} are based on BERT~\citep{bert}, while more recent models such as Qwen3-Embedding~\citep{qwen3_embedding} achieve significantly improved retrieval performance.
\textbf{Structure-Augmented RAG}: These approaches enhance retrieval with structural representations such as graphs or trees. Representative systems include GraphRAG~\citep{graphrag}, RAPTOR~\citep{sarthi2024raptor}, HippoRAG-V2~\citep{hipporag-v2}, Cognee, Zep~\citep{Zep}, MemoRAG~\citep{memorag}, Mem0~\citep{Mem0}, MemoryOS~\citep{memoryos}, Memary~\citep{memary} and Memobase~\citep{memobase}. Despite their effectiveness, RAG-based methods face challenges with ambiguous queries, multi-hop reasoning, and long-range comprehension. When questions require integrating knowledge across an entire session or learning from long, skill-encoding inputs, the retrieval mechanism—limited to the top-k most relevant passages—may fail to surface the necessary information. 
To address these limitations, \textbf{Agentic Memory Agents} introduce an iterative, decision-driven framework. Rather than relying on a single-pass retrieval, these agents dynamically process the query, retrieve evidence, reflect, and iterate through multiple retrieval and reasoning cycles. Examples include MemGPT~\citep{MemGPT}, Self-RAG~\citep{self-rag}, Auto-RAG~\citep{auto-rag}, A-MEM~\citep{a-mem}, Mem1~\citep{mem1}, MemAgent~\citep{MemAgent}, and MIRIX~\citep{mirix}. This agentic design is particularly effective for resolving ambiguous or multi-step queries. Nonetheless, these methods remain fundamentally constrained by the limitations of RAG—namely, the inability to fully understand or learn from long-range context that is inaccessible via retrieval alone.
\vspace{-6pt}

%% file: sections/method.tex
\begin{table*}[tbh]
    \centering
    \caption{
            Overview of evaluation datasets. 
            We select datasets that cover various important long-context capabilities. In the table, we underline the datasets we constructed ourselves. AvgL.: Average Context Length (measured using the GPT-4o-mini model's tokenizer).
    }
    \resizebox{\textwidth}{!}{
        \begin{tabular}{lllll}
            \toprule
            \textbf{Category} & \textbf{Dataset}   & \textbf{Metrics} & \textbf{AvgL.} & \textbf{Description} \\

            \midrule
            \multirow{4}{7em}{\textbf{Accurate Retrieval}} 
            & SH-Doc QA         & \multirow{4}{*}{Accuracy} & 197K & Single-Hop Gold passage retrieval QA.  \\
            & MH-Doc QA         &  & 421K & Multiple-Hop Gold passage retrieval QA.  \\
            & LongMemEval (S*)  &  & 355K & Dialogues based QA.  \\
            & EventQA           &  & 534K & Novel multiple-choice QA on characters events. \\

            \midrule
            \multirow{6}{7em}{\textbf{Test-time Learning}} & BANKING77  &  \multirow{5}{*}{Accuracy} & \multirow{5}{*}{103K} & Banking intent classification, 77 labels. \\
            & CLINC150 &  &  & Intent classification, 151 labels. \\
            & NLU &  &  & Task intent classification, 68 labels. \\
            & TREC Coarse &  &  & Question type classification, 6 labels. \\
            & TREC Fine &  &  & Question type classification, 50 labels. \\
            & Movie Recommendation  & Recall@5 & 1.44M & Recommend movies based on provided dialogues examples. \\

             \midrule
            \multirow{2}{8em}{\textbf{Long Range Understanding}} 
            & $\infty$Bench-Sum & F1-Score   & 172K & Novel summarization with entity replacement.   \\
            & Detective QA      & Accuaracy        & 124K & Long-range reasoning QA on detective novels.   \\

            \midrule
            \multirow{2}{7em}{\textbf{Selective Forgetting}}  & FactConsolidation-SH   & \multirow{2}{*}{Accuracy}     & \multirow{2}{*}{262K} &   Single hop reasoning in facts judgment.  \\
            & FactConsolidation-MH  &      &  & Multiple hop reasoning in facts judgment. \\
            \bottomrule
        \end{tabular}
        }
    \label{tab:dataset_overview_details}
\end{table*}

\section{\ours}

\subsection{Dataset Preperation}
In this section, we describe how we reconstruct existing datasets and build new ones for evaluating each competency aspect. All datasets with their categories are shown in Table \ref{tab:dataset_overview_details}. We introduce the details in datasets curation in Appendix~\ref{sec:details_of_datasets}. 

\vspace{-8pt}
\paragraph{Datasets for Accurate Retrieval (AR)}
We adopt four datasets to evaluate the accurate retrieval capability of memory agents. Three are reconstructed from existing benchmarks, and one is newly created: (1) \textbf{Document Question Answering}: This is a NIAH-style QA task where a long passage contains single (SH-QA) or multiple (MH-QA) documents answering the input question. The agent must identify and extract relevant snippets from the extended context. (2) \textbf{LongMemEval}: This benchmark evaluates memory agents on long dialogue histories. Although task types like information extraction (IE) or multi-session reasoning are included, most tasks can be reformulated as single-retrieval problems requiring agents to retrieve the correct segments spanning a long multi-turn conversation. We reformulated chat history into five long dialogues ($\sim$355K tokens) with 300 questions (LongMemEval (S*) in Table ~\ref{tab:dataset_overview_details}). We create LongMemEval (S*) specifically for increasing the number of questions per context, mitigating the exhaustive needs of reconstructing the memory for each question. (3) \textbf{EventQA (ours)}: We introduce EventQA this reasoning style NIAH task to evaluate agents’ ability to recall and reason about temporal sequences in long-form narratives. In this dataset, the agent is required to read a novel and select the correct event from a series of candidates after receiving up-to five previous events. Unlike other long-range narrative text datasets that require extensive manual annotation~\citep{infinitebench, detectiveqa}, our dataset is built through a fully automated pipeline, making the process more efficient and scalable. Moreover, this pipeline can be directly applied to other novel-style texts.

\vspace{-8pt}

\paragraph{Datasets for Test-Time Learning (TTL)}
We evaluate TTL via two task categories: (1) \textbf{Multi-Class Classification (MCC)}: We reconstructed five classification datasets used in prior TTL work~\citep{icl,HELMET}: BANKING77~\citep{banking77}, CLINC150~\citep{clinic150}, TREC-Coarse, TREC-Fine~\citep{trec}, and NLU~\citep{nlu}. Each task requires the agent to map sentences to class labels, leveraging previously seen labeled examples in context. (2) \textbf{Recommendation}: Based on the setup from \citep{redial, LLM_Zeroshot_Recsys}, we construct a dataset to evaluate movie recommendation via dialogue history. The agent is exposed to thousands of movie-related dialogue turns and is asked to recommend twenty relevant movies based on the long interaction history.

\vspace{-8pt}
\paragraph{Datasets for Long Range Understanding (LRU)} We evaluate LRU via two tasks: (1) \textbf{Novel Summarization (Summ.)}: We adopt the Summarization task \texttt{En.Sum} from $\infty$-Bench~\citep{infinitebench}. The agent is required to analyze and organize the plot and characters of the novel, and then compose a summary of 1000 to 1200 words. (2) \textbf{Detective QA (Det QA)}: We also create a difficult question set from Detective QA~\citep{detectiveqa}, which include ten novels with 71 questions and these questions require agents to do reasoning over a longer narrative range.

\vspace{-8pt}
\paragraph{Datasets for Selective Forgetting (SF)}
To assess whether an agent can forget out of date memory and reason over them, we construct a new dataset called FactConsolidation. Specifically, We build this benchmark using counterfactual edit pairs from \textsc{MQUAKE}~\citep{mquake}. Each pair contains a true fact and a rewritten, contradictory version. These are ordered such that the rewritten (new) fact appears after the original, simulating a realistic update scenario. We concatenate multiple such edit pairs to create long contexts of length {6K, 32K, 64K, 262K}. We then adpot MQUAKE’s original questions and categorize them into: (1) \textbf{FactConsolidation-SH (Ours)} (SH means Single-Hop), requiring direct factual recall (e.g., “Which country was tool $A$ created in?”), and (2) \textbf{FactConsolidation-MH (Ours)} (MH refers to Multi-Hop), requiring inference over multiple facts (e.g., “What is the location of death of the spouse of person $B$?”). 
Agents are prompted to prioritize later information in case of conflict and reason based on the final memory state. This setup directly evaluates the strength and consistency of selective forgetting over long sequences.

\subsection{Different Categories of Memory Agents}
\label{sub:various_memory_agents}

We evaluate three major types of memory agents that reflect common strategies for handling long-term information: \emph{Long-Context Agents}, \emph{RAG Agents}, and \emph{Agentic Memory Agents}. These approaches differ in how they store, retrieve, and reason over past inputs. 

\textbf{(1) Long Context Agents}
Modern language models often support extended context windows ranging from 128K to over 1M tokens. A straightforward strategy for memory is to maintain a context buffer of the most recent tokens. For example, in a model with a 128K-token limit, the agent concatenates all incoming chunks until the total exceeds the window size. Once the limit is reached, the earliest chunks are evicted in a FIFO (first-in, first-out) manner. This agent design relies solely on positional recency and assumes the model can attend effectively over the current context window. \textbf{(2) RAG Agents}
RAG-based agents address context limitations by storing past information in an external memory pool and retrieving relevant content as needed. We consider three RAG variants: \emph{Simple RAG Agents}: All input chunks are stored as raw text. During inference, a keyword or rule-based string matching mechanism retrieves relevant passages. \emph{Embedding-based RAG Agents}: Each input chunk is embedded and saved. At query time, the agent embeds the query and performs retrieval using cosine similarity between embeddings. \emph{Structure-Augmented RAG Agents}: After ingesting all input chunks, the agent constructs a structured representation (e.g., knowledge graph or event timeline). Subsequent queries are answered based on this structured memory. \textbf{(3) Agentic Memory Agents}
Agentic memory agents extend beyond static memory stores by employing agentic loops—iterative reasoning cycles in which the agent may reformulate questions, perform memory lookups, and update its working memory. These agents are designed to simulate a more human-like process of recalling, verifying, and integrating knowledge.

\subsection{Datasets and Agents Formulation}

\paragraph{Datasets Formulation}
We standardize all datasets into the format: ${c_1, c_2, \cdots, c_n}$ (chunks), ${q_1, q_2, \cdots, q_m}$ (questions), and ${a_1, a_2, \cdots, a_m}$ (answers), where $c_i$ denotes the $i$-th chunk wrapped to construct a user message with instructions of memorizing the content in a sequential input, and ${c_1, c_2, \cdots, c_n}$ represents a single conversation. Each chunk is accompanied by instructions prompting the agent to memorize its contents. Example prompts are provided in Appendix~\ref{sub:instructions_for_memory_construction}. When curating datasets like EventQA and FactConsolidation, we deliberately design scenarios where multiple questions follow a single context. This allows us to probe the model’s memory multiple times with one sequential injection. For example, in LME (S*), five contexts are paired with 300 questions (shown in Table \ref{tab:datasets} in Appendix \ref{sec:details_of_datasets}). This design choice reflects a key trend: as LLMs support increasingly long context windows and memory agents become more capable of handling extended inputs, evaluation datasets must also scale accordingly. Injecting 1M tokens for just one question is resource-inefficient, whereas associating the same input with many questions provides significantly higher utility.

\paragraph{Prompt Formulation and Interaction Protocol} Unlike standard long-context evaluations that input raw text, we wrap all input chunks within a simulated User-Assistant dialogue to explicitly trigger the agent's memory mechanism. Each input chunk $c_i$ is preceded by a memorization instruction (e.g., ``Please memorize it and I will ask some questions...'') to establish a clear intent for information storage. Simultaneously, for each specific dataset, we carefully curated the instructions to ensure agents accurately comprehend the task intent and execute the required actions.  Crucially, for the Selective Forgetting competency, we introduce explicit guardrails in the prompt. We explicitly instruct agents that facts are indexed by serial numbers, and that ``\textit{newer facts have larger serial numbers.}''. The agents are mandated to solve conflicts by finding the newest fact (see Appendix ~\ref{sec:prompts} for full templates).

\paragraph{Agents Formulation} 
In our framework, all agents are required to take the chunks one by one, absorb them into memory, and incrementally update the memory. After seeing all the chunks, we ask the agent to answer the related questions. To guarantee fair comparison, we employed standardized prompt templates across all agents within each evaluation category, with only minimal adaptations where necessary. 



%% file: sections/experiments.tex
\section{Experiments}

\subsection{Experimental Setup}
The datasets are split into four categories and the statistics of all datasets are also shown in Table \ref{tab:datasets}. The evaluation metrics for all datasets are shown in Table \ref{tab:dataset_overview_details}, along with more dataset details. For the agents, as described in Section \ref{sub:various_memory_agents}, we consider three categories of agents: \emph{Long-Context Agents}, \emph{RAG agents} and \emph{Agentic Memory Agents}, where \emph{RAG Agents} can be further split into \emph{Simple RAG Agents}, \emph{Embedding-based RAG Agents} and \emph{Structure-Augmented RAG Agents}. We give the detailed introduction of each memory agent in Appendix ~\ref{sec:agents_intro}.  For chunk size settings, we choose a chunk size of 512 for the SH-Doc QA, MH-Doc QA, and LME(S*) tasks in AR, as well as for all tasks in SF. This is mainly because these tasks are composed of long texts synthesized from multiple short texts. For other tasks, we use a chunk size of 4096. Considering computational overhead and API cost, we uniformly use a chunk size of 4096 for Mem0, Cognee, Zep, and MIRIX. We report the detailed settings of the chunk size in Table ~\ref{tab:different_context_length_setting} in Appendix ~\ref{sec:detailed_exp_results}. 
\vspace{-6pt}

\begin{table*}[t]
    \centering
    \setlength{\tabcolsep}{2pt} 
    \caption{Overall Performance Comparison. In the absence of a specified model, All RAG agents and commercial memory agents use GPT-4o-mini as the backbone. Thus we highlight the performance of GPT-4o-mini as the reference. FC-SH and FC-MH mean FactConsolidation Single Hop and FactConsolidation Multi Hop, respectively. Best viewed in colors.} 
    \resizebox{\textwidth}{!}{
    \begin{tabular}{l|cccc>{\columncolor{avgcolor}}c|cc>{\columncolor{avgcolor}}c|cc>{\columncolor{avgcolor}}c|cc>{\columncolor{avgcolor}}c|>{\columncolor{overallcolor}}c}
        \rowcolor{headergray}
        & \multicolumn{5}{c}{\textbf{AR}} & \multicolumn{3}{c}{\textbf{TTL}} & \multicolumn{3}{c}{\textbf{LRU}} & \multicolumn{3}{c}{\textbf{SF}} & \cellcolor{overallcolor}Overall \\
        \rowcolor{headergray}
        \textbf{Agent Type} & SH-QA & MH-QA & LME(S*) & EventQA & \cellcolor{avgcolor}\textbf{Avg.} & MCC & Recom. & \cellcolor{avgcolor}\textbf{Avg.} & Summ. & DetQA & \cellcolor{avgcolor}\textbf{Avg.} & FC-SH & FC-MH & \cellcolor{avgcolor}\textbf{Avg.} & \cellcolor{overallcolor}Scores \\
        \toprule
        \rowcolor{groupblue}
        \multicolumn{16}{c}{\emph{Long-Context Agents}} \\
        GPT-4o (128K)           & 72.0 & 51.0 & 32.0 & 77.2 & 58.1 & 87.6 & 12.3 & 50.0 & 32.2 & 77.5 & 54.9 & 60.0 & 5.0 & 32.5 & 48.8  \\
        GPT-4o-mini (128K)      & 64.0 & 43.0 & 30.7 & 59.0 & 49.2 & 82.0 & 15.1 & 48.6 & 28.9 & 63.4 & 46.2  & 45.0 & 5.0 & 25.0 & 42.2  \\
        
        Claude-3.7-Sonnet (200K) & 77.0 & 53.0 & 34.0 & 74.6 & 59.7 & \textbf{89.4} & \textbf{18.3} & \textbf{53.9}  & 52.5 & 71.8 & 62.2 & 43.0 & 2.0 & 22.5 & 49.6 \\
        
        GPT-5-mini (400K) & 85.0 & \textbf{71.0} & \textbf{63.3} & \textbf{78.2} & \textbf{74.4} & 84.0 & 13.2 & 48.6 & \textbf{56.3} & \textbf{76.1} & \textbf{66.2} & 78.0 & 28.0 & \textbf{53.0} & \textbf{60.6} \\

        GPT-4.1-mini (1M)     & 83.0 & 66.0 & 55.7 & 82.6 & 71.8 & 75.6 & 16.7 & 46.2 & 41.9 & 56.3 & 49.1 & 36.0 & 5.0 & 20.5 & 46.9 \\

        Gemini-2.0-Flash (1M)  & \textbf{87.0} & 59.0 & 47.0 & 67.2 & 65.1 & 84.0 & 8.7 & 46.4 & 23.9 & 59.2 & 41.6 & 30.0 & 3.0 & 16.5 & 42.4   \\

        \midrule
        \midrule
        \rowcolor{headergray}
        GPT-4o-mini & 64.0 & 43.0 & 30.7 & 59.0 & \cellcolor{avgcolor}49.2 & 82.0 & 15.1 & \cellcolor{avgcolor}48.6 & 28.9 & 63.4 & \cellcolor{avgcolor}46.2 & 45.0 & 5.0 & \cellcolor{avgcolor}25.0 & \cellcolor{overallcolor}42.3 \\
        \rowcolor{groupgreen}
        \multicolumn{16}{c}{\emph{Simple RAG Agents}} \\
        BM25              & 66.0 & 56.0 & 45.3 & \textbf{74.6} & 60.5 & 75.4 & 13.6 & 44.5 & \textbf{19.0} & 52.1 & 35.6 & \underline{48.0} & 3.0 & \underline{25.5} & \underline{41.5}  \\
        \rowcolor{groupgreen}
        \multicolumn{16}{c}{\emph{Embedding RAG Agents}} \\
        Contriever         & 22.0 & 31.0 & 15.7 & 66.8 & 33.9 & 70.6 & 15.2 & 42.9 & 17.2 & 42.3 & 29.8 & 18.0 & \textbf{7.0} & 12.5 & 29.8 \\
        Text-Embed-3-Small & 60.0 & 44.0 & 48.3 & 63.0 & 53.8 & 70.0 & \underline{15.3} & 42.7 & 17.7 & 54.9 & 36.3 & 28.0 & 3.0 & 15.5 & 37.1  \\
        Text-Embed-3-Large & 54.0 & 44.0 & 50.3 & 70.0 & 54.6 & 72.4 & \textbf{16.2} & 44.3 & 18.2 & 56.3 & \underline{37.3} & 28.0 & 4.0 & 16.0 & 38.0 \\
        Qwen3-Embedding-4B & 57.0 & 47.0 & 43.3 & \underline{71.4} & 54.7 & \textbf{78.0} & 12.2 & \textbf{45.1} & 14.8 & \underline{59.2} & 37.0 & 29.0 & 3.0 & 16.0 & 38.2 \\

        \rowcolor{groupgreen}
        \multicolumn{16}{c}{\emph{Structure-Augmented RAG Agents}} \\
        RAPTOR      & 29.0 & 38.0 & 34.3 & 45.8 & 36.8 & 59.4 & 12.3 & 35.9 & 13.4 & 42.3 & 27.9 & 14.0 & 1.0 & 7.5 & 27.0 \\
        GraphRAG    & 47.0 & 47.0 & 35.0 & 34.4 & 40.9 & 39.8 & 9.8 & 24.8 & 0.4 & 39.4 & 19.9 & 14.0 & 2.0 & 8.0 & 23.4 \\
        MemoRAG     & 29.0 & 33.0 & 20.0 & 56.0 & 34.5 & \underline{77.0} & 13.1 & \textbf{45.1} & 9.2 & 50.7 & 30.0 & 21.0 & \textbf{7.0} & 14.0 & 30.9 \\
        HippoRAG-v2 & \textbf{76.0} & \underline{66.0} & \underline{50.7} & 67.6 & \textbf{65.1} & 61.4 & 10.2 & 35.8  & 14.6 & 57.7 & 36.2 & \textbf{54.0} & 5.0 & \textbf{29.5} & \textbf{41.6} \\
        Mem0        & 25.0 & 32.0 & 36.0 & 37.5 & 32.6 & 32.4 & 10.0 & 21.2 & 4.8 & 36.6 & 20.7 & 18.0 & 2.0 & 10.0 & 21.1 \\
        Cognee      & 31.0 & 26.0 & 29.3 & 26.8 & 28.3 & 35.4 & 10.1 & 22.8 & 2.3 & 29.6 & 16.0 & 28.0 & 3.0 & 15.5 & 20.6 \\
        Zep            & 44.0 & 25.0 & 38.3 & 42.5 & 37.5 & 62.8 & 12.1 & 37.5 & 4.2 & 28.2 & 16.2 & 7.0 & 3.0 & 5.0 & 24.0 \\
        \rowcolor{grouppink}
        \multicolumn{16}{c}{\emph{Agentic Memory Agents}} \\
        Self-RAG       & 35.0 & 42.0 & 25.7 & 31.8 & 33.6 & 11.6 & 12.8 & 12.2 & 0.9 & 35.2 & 18.1 & 19.0 & 3.0 & 11.0 & 18.7 \\
        MemGPT         & 41.0 & 38.0 & 32.0 & 26.2 & 34.3 & 67.6 & 14.0 & 40.8 & 2.5 & 42.3 & 22.4 & 28.0 & 3.0 & 15.5 & 28.3 \\
        MIRIX           & 62.0 & 61.0 & 37.3 & 29.8 & 47.5 & 38.4 & 9.8 & 24.1 & 9.9 & 40.8 & 25.4 & 14.0 & 2.0 & 8.0 & 26.2 \\
        MIRIX (4.1-mini)  & \underline{73.0} & \textbf{75.0} & \textbf{51.0} & 53.0 & \underline{63.0} & 61.0 & 10.3 & 35.7 & \underline{18.9} & \textbf{62.0} & \textbf{40.5} & 20.0 & 3.0 & 11.5 & 37.7 \\
        \bottomrule
    \end{tabular}
    }
    \vspace{-8pt}
\label{tab:overall_performance_comparison}
\end{table*}

\subsection{Overall Performance Comparison}
Table~\ref{tab:overall_performance_comparison} presents the overall performance across different benchmarks. We summarize the key findings as follows: \textbf{(1) Superiority of RAG methods in Accurate Retrieval Tasks.} Most RAG Agents are better than the backbone model ``GPT-4o-mini'' in the tasks within the Accurate Retrieval Category. This matches our intuition where RAG agents typically excel at extracting a small snippet of text that is crucial for answering the question. \textbf{(2) Superiority of Long-Context Models in Test-Time Learning and Long-Range Understanding.} Long-context models achieve the best performance on TTL and LRU. This highlights a fundamental limitation of RAG methods and commercial memory agents, which still follow an agentic RAG paradigm. These systems retrieve only partial information from the past context, lacking the ability to capture a holistic understanding of the input—let alone perform learning across it. \textbf{(3) Limitation of All Existing Methods on Selective Forgetting.} Although being a well-discussed task in model-editing community~\citep{SERAC,alphaedit}, forgetting out-of-date memory poses a significant challenge on memory agents. We observe that all methods fail on the multi-hop situation (with achieving at most 28\% accuracy). Only long context agents can achieve fairly reasonable results on single-hop scenarios. In Section \ref{sec:factcon_validation}, we show that current reasoning models can have much better performance, while it does not change the conclusion that Selective Forgetting still poses a significant challenge to all memory mechanisms.

\subsection{Analysis and Ablation Study}

In this section, we present experiments and analysis along five dimensions: input chunk size, retrieval top-$k$, backbone model, and dataset validation. Additional results are provided in the appendix, including compuational latency (Appendix \ref{sub:latency_analysis}), context length analysis (Appendix \ref{sub:different_context_length}), latency and GPU memory usage comparisons (Appendix \ref{sub:latency_analysis}, \ref{sub:gpu_memory_usage_comparison}),  further details on chunk size and top-$k$ ablations (Appendix \ref{sub:input_chunk_ablation}, \ref{sub:results_on_topk}), as well as the Cost-Performance estimation (Appendix \ref{app:cost_performance}).

\subsubsection{Ablation Study on Input Chunk Size}

\begin{figure}[t]
    \centering
    \begin{subfigure}[b]{0.495\linewidth}
        \centering
        \includegraphics[width=\linewidth]{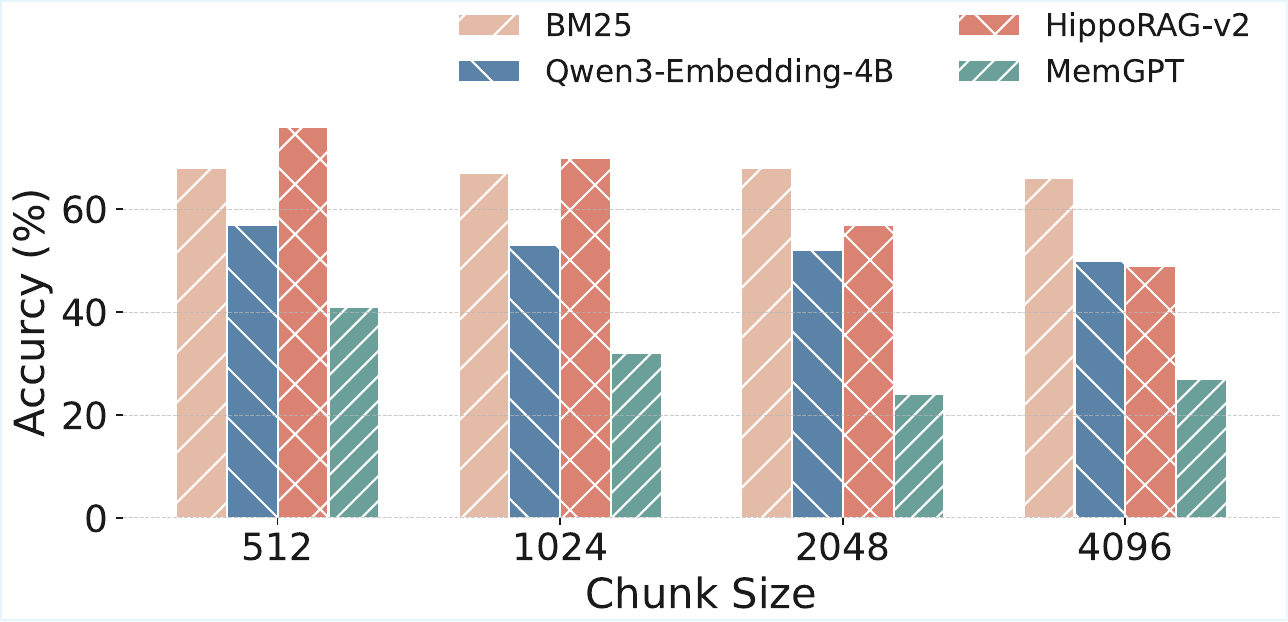}
        \caption{SH-Doc QA performance}
        \label{fig:ruler_chunk}
    \end{subfigure}
    \hfill
    \begin{subfigure}[b]{0.495\linewidth}
        \centering
        \includegraphics[width=\linewidth]{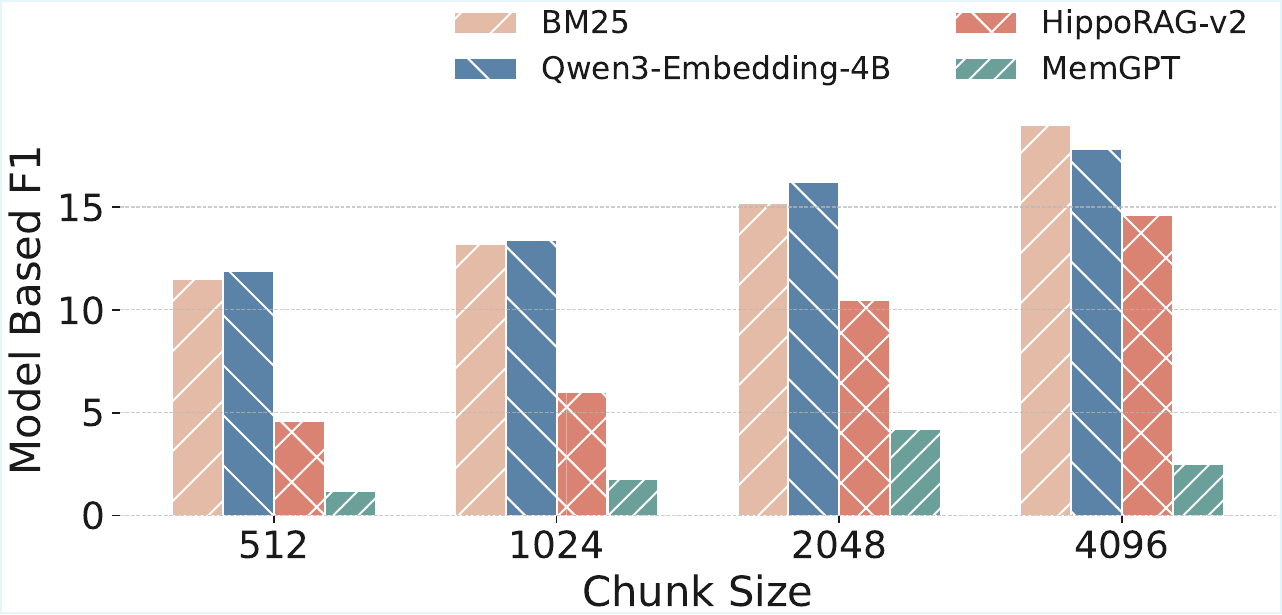}
        \caption{$\infty$Bench-Sum performance}
        \label{fig:infinite_sum_chunk}
    \end{subfigure}
    \caption{Performances on SH-Doc QA and $\infty$-Bench-Sum with different chunk sizes.}
    \label{fig:ruler_chunk_size}
    \vspace{-10pt}
\end{figure}

To understand how chunk size impacts performance, particularly for RAG methods and agentic memory agents, we conduct an additional analysis where we vary the chunk size while fixing the number of retrieved chunks to 10. 
The results are presented in Figure~\ref{fig:ruler_chunk_size}.
From the figure, we observe 
that when resources permit, using smaller chunk sizes and increasing the number of retrieval calls during memory construction can improve performance on Accurate Retrieval (AR) tasks. Finer-grained segmentation enhances the relevance of retrieved information, particularly for embedding-based methods. However, for tasks requiring Long-Range Understanding (LRU), varying the chunk size hurts the performance. This is likely because RAG methods are inherently less suited for tasks that demand integration of information across a large, coherent context.

\subsubsection{Ablation Study on Retrieval TopK}
In our experiments, although we report most results with the number of retrieved chunks set to 10 in Table~\ref{tab:overall_performance_comparison}, we also conducted ablation studies with varying retrieval sizes. A subset of these results is visualized in Figure~\ref{fig:retrieve_topk_ablation}, with the full results provided in Table~\ref{tab:retrieval_topk_ablation} in Appendix ~\ref{sec:detailed_exp_results}.
The results indicate that increasing the number of retrieved chunks generally improves performance across most tasks. It is worth noting that, with a chunk size of 4096 tokens, retrieving 10 chunks already yields an input of approximately 40k tokens. This places significant demands on model capacity. Due to this high token volume, we do not evaluate settings with 20 retrieved chunks.

\begin{figure}
\centering
\includegraphics[width=\linewidth]{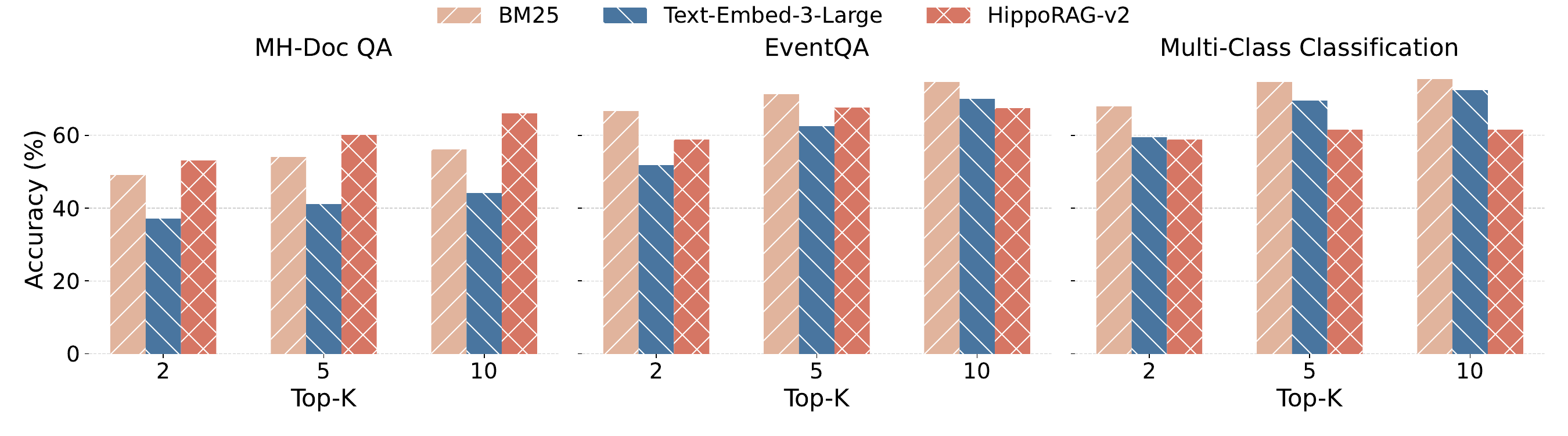}
\caption{The accuracies on different benchmarks when varying the retrieval top-k to be 2, 5 and 10.}
\label{fig:retrieve_topk_ablation}
\vspace{-10pt}
\end{figure}

\subsubsection{Ablation Study on Backbone Model}

To investigate how different backbone models impact the performance of various memory agents, we experimented with three different backbone models and selected four representative methods from both the RAG Agents and Agentic Memory categories. The complete experimental results are presented in Table ~\ref{tab:different_backbone}. Our findings show that for RAG Agents, once the backbone is sufficiently strong, it no longer serves as the main performance bottleneck. Compared to the default setup, upgrading to a more powerful model like GPT-4.1-mini yields only marginal improvements. In contrast, the main results in Table ~\ref{tab:overall_performance_comparison} for the MIRIX method under the Agentic Memory category, using a stronger backbone leads to substantial performance gains. This suggests that future advances in backbone models could further boost the effectiveness of Agentic Memory methods.

\begin{table*}
    \centering
    \caption{Performance comparison on three different backbone LLMs and four representative memory agents. We choose one dataset from every competency to evaluate agent performance. }
    \resizebox{\textwidth}{!}{
    \begin{tabular}{cc|ccccc}
        \toprule
          \rowcolor{headergray}
          \textbf{Agent Type} & Backbone Model  & EventQA & Recom & $\infty$Bench-Sum & FactCon-SH & Avg. \\
          \midrule
            & GPT-4o-mini      & 74.6 & 13.6 & 19.0 & 48.0 & 38.8  \\
          BM25  & GPT-4.1-mini     & 76.4 & 14.0 & 19.4 & 51.0 & 40.2  \\
            & Gemini-2.0-Flash & 70.8 & 10.0 & 18.9 & 47.0 & 36.7 \\
          \midrule
            & GPT-4o-mini       & 63.0 & 15.3 & 17.7 & 28.0  & 31.0 \\
          Text-Embed-3-Small & GPT-4.1-mini      & 62.0 & 15.5 & 17.9 & 30.0 & 31.4  \\
            & Gemini-2.0-Flash  & 64.0 & 10.3 & 17.2 & 27.0 & 29.6 \\
          \midrule
            & GPT-4o-mini        & 34.4 & 9.8 & 0.4 & 14.0 & 14.7   \\
          GraphRAG & GPT-4.1-mini        & 39.0 & 10.3 & 1.2 & 16.0 & 16.6   \\
            & Gemini-2.0-Flash     & 36.2 & 7.2 & 0.8 & 13.0 & 14.3   \\
          \midrule
          \multirow{2}{*}{MIRIX}  & GPT-4o-mini  & 29.8 & 9.8 & 9.9 & 14.0 & 15.9  \\
          & GPT-4.1-mini  & 53.0 \textbf{(23.2$\uparrow$)} & 10.3 \textbf{(0.5$\uparrow$)} & 18.9 \textbf{(9.0$\uparrow$)} & 20.0 \textbf{(6.0$\uparrow$)} & 25.6 \textbf{(9.7$\uparrow$)}   \\
          \bottomrule
        \end{tabular}}
    \label{tab:different_backbone}
     \centering
     
\end{table*}

\subsubsection{Validation of Dataset FactConsolidation}
\label{sec:factcon_validation}

\begin{wraptable}{r}{0.48\textwidth} 
    \centering
    \vspace{-12pt}
    \caption{Performances of reasoning models on the dataset FactConsolidation.}
      \resizebox{0.4\textwidth}{!}{
    \begin{tabular}{c|cc|cc}
        \toprule
          & \multicolumn{2}{c|}{FactCon-SH} & \multicolumn{2}{c}{FactCon-MH} \\
         & 6K & 32K & 6K & 32K \\
        \midrule
        GPT-4o   & 92.0 & 88.0 & 28.0 & 10.0 \\
        O4-mini  & \textbf{100.0} & 61.0 & \textbf{80.0} & 14.0 \\
        \bottomrule
    \end{tabular}
    }
\label{tab:factconsolidation_check}
\vspace{-10pt}
\end{wraptable}

As the performance of different models on this dataset
remains drastically low, we turn to the stronger reasoning model o4-mini and validate our dataset by checking the performance of o4-mini on a smaller version
of this dataset. The results are shown in Table ~\ref{tab:factconsolidation_check}.  We found that on the 6K version of the FactCon-SH dataset, both models perform well and are generally able to complete the task effectively. However, their performance drops when the context length increases to 32K. Similarly, on the 6K version of the FactCon-MH dataset, the stronger O4-mini reasoning model achieves a decent score of 80.0, but its performance significantly drops to 14.0 when the context window reaches 32K. This indicates that our dataset is solvable under short-context settings, but current memory agents still lack strong long-range reasoning capabilities, making them unable to handle the task when presented with longer historical inputs.


%% file: sections/conclusion.tex
\section{Conclusion}
In this paper, we introduce \textbf{\ours}, a unified benchmark designed to evaluate memory agents across four essential competencies: accurate retrieval, test-time learning, long-range understanding, and selective forgetting. While prior benchmarks focus largely on skill execution or long-context question answering, \ours fills a critical gap by assessing how agents store, update, and utilize long-term information across multi-turn interactions. To build this benchmark, we restructure existing datasets and propose two new ones—\textbf{EventQA} and \textbf{FactConsolidation}—tailored to stress specific memory behaviors often overlooked in prior work. We evaluate a wide spectrum of agents, including long-context models, RAG-based systems, and commercial memory agents, under a consistent evaluation protocol. Our results reveal that, despite recent advances, current memory agents still exhibit substantial limitations when faced with tasks requiring dynamic memory updates and long-range consistency. 
One limitation of our work is that due to budget constraints, so we could only conduct experiments on some relatively representative Memory Agents. 
As future work, we aim to provide more evaluation results for more memory agents.

%% file: sections/statement.tex
\section*{Ethics Statement}

This work adheres to the ICLR Code of Ethics and associated author guidance; we assess potential impacts and document mitigations accordingly. We evaluate memory in LLM agents using dialogs , license-compliant corpora; no personally identifiable information or data from minors were collected. To reduce dual-use risks, we release only safety-screened prompts and provide usage notes discouraging surveillance-oriented applications. We will release code under the MIT License and datasets/benchmark artifacts under CC BY 4.0; third-party materials retain their original licenses.

\section*{Reproducibility Statement}

Upon acceptance, we will open-source \textbf{all code and data} used in this paper. The repository will include (i) training/evaluation scripts, configuration files, and exact prompts; (ii) dataset releases and generation scripts with seeds to fully regenerate  interactions; and (iii) end-to-end run recipes. We will pin software dependencies and provide a containerized environment (Dockerfile plus conda/requirements.txt) and report hardware, CUDA/cuDNN, and OS details to support deterministic re-runs, in line with community guidance on reproducibility statements and artifact preparation. 

%% file: sections/appendix.tex
\newpage
\appendix

\section{The Use of Large Language Models (LLMs)}

In this paper writing process, we used an LLM to assist with content polishing—for example, identifying grammatical errors and suggesting revisions for sentences that were unclear or potentially ambiguous. Additionally, we used the LLM to generate character icons, which were later used in the creation of our main plot visualization.

\section{Details of Dataset}
\label{sec:details_of_datasets}

Here we provide a detailed introduction to the datasets used for evaluating the four core competencies, including the dataset curation, corresponding metrics, average context length, and a brief description. Details are shown in Table \ref{tab:dataset_overview_details}.

\subsection{Accurate Retrieval (AR)}

\subsubsection{Definition of AR}
The task of accurately retrieving information has been extensively explored in prior work. In the domain of long-context modeling, the Needle-in-a-Haystack (NIAH) task is widely used to evaluate a model’s ability to locate the specific value based on a given key within a lengthy input. In the RAG setting, this corresponds to document-based QA, where the model must identify and extract relevant snippets from one or more documents to answer a query. These snippets might reside in a single location or be distributed across multiple documents.
In this paper, we focus on agentic settings, where the “long context” or “multiple documents” become long-form conversations. We define Accurate Retrieval (AR) as the ability of an agent to identify and retrieve important information that may be dispersed throughout a long dialogue history. 

\subsubsection{Details on AR datasets}
We use four datasets to evaluate the accurate retrieval capability of memory agents. 
\paragraph{(1) Document Question Answering} We improved two QA datasets from~\citep{RULER}. These datasets provide multiple synthetic contexts of varying lengths, ranging from 3K to over 200K tokens. We select 100 questions from the datasets with shorter context length. For each of these 100 questions, we collect the context and remove duplicate short documents, and then shuffle and concatenate them to create new long documents of 197K or 421K tokens, making sure the new context containing the gold passages. Since most answers are short informational entities, such as years, names, or yes/no responses, we use substring exact match (SubEM) to calculate the accuracy of QA. SubEM measures whether the predicted answer exactly matches the gold answer as a sub-string, which is a common standard in question answering systems. 

\paragraph{(2) LongMemEval} This is a dialogue-based QA dataset. For LME(S*), we use multiple historical conversation data segments, arrange them in chronological order, and finally concatenate them to create five long conversation histories, each with a length of approximately 355K tokens. Since some of the questions have open-ended answers, we adopt the approach used in previous work and employ the GPT-4o model to assess whether the agent's responses meet the requirements. If a response is deemed satisfactory, it is marked as True. Finally, we calculate the proportion of satisfactory responses as the evaluation metric. ~\citet{longmemeval} reported in Table 6 that a prompt-engineered GPT-4o judge achieves 98.0\% accuracy and demonstrates very high stability.

\paragraph{(3) EventQA} Using five books from $\infty$-Bench (each contains more than 390K tokens, counted using the \texttt{gpt-4o-mini} tokenizer), we identify the ten most frequently mentioned characters via \texttt{SpaCy} NER. We extract 101 events experienced by key characters using \texttt{gpt-4o}. For each event, we construct a 6-way multiple-choice question by pairing the true event with five distractors generated via \texttt{gpt-4o}. The agent receives up-to five previous events and must identify the correct continuation. We report the mean accuracy over 100 such questions per book, and ultimately present the average accuracy across all five books.

\subsection{Test-time Learning (TTL)}

\subsubsection{Definition of TTL}

An essential capability for real-world agents is the ability to acquire new skills dynamically through interaction with users. This mirrors the concept of In-Context Learning (ICL) in LLMs, where the model learns from a prompt containing a small number of examples, often framed as few-shot classification tasks. Ideally, performance improves with additional examples in the prompt.
In the conversational agent setting, prompts are replaced by dialogue histories. We define Test-Time Learning (TTL) as the agent’s ability to learn to perform new tasks directly from the conversation. This property is crucial for enabling self-evolving agents that can continuously adapt and improve in real-world deployments.

\subsubsection{Details on TTL datasets}

We evaluate TTL via two task categories:

\paragraph{(1) Multi-Class Classification (MCC)} We adopt five classification datasets used in prior TTL work. For dataset curation, we use thousands of sentence samples from different categories, with each type of sample assigned a number as its label. Following the format "\{sentence\} {\textbackslash n} Label: \{label\} {\textbackslash n}", we concatenate all the sentences into a long context and shuffle them to prevent samples of the same type from being too concentrated. In this task, the agent needs to refer to a long context and correctly classify the input content. Therefore, we use average accuracy as the evaluation metric. 

\paragraph{(2) Recommendation (Recom.)} We concatenate multiple short dialogues about movie recommendations from the original dataset, remove duplicate dialogues, and create a long context containing over a thousand recommendation instances. In this task, the agent is required to recommend 20 movies based on the content of the dialogue. We evaluate the recommendations by calculating Recall@5, which measures the overlap between the top 5 recommended movies and the ground truth.

\subsection{Long-Range Understanding (LRU)}

\subsubsection{Definition of LRU}
Long-range understanding refers to the agent’s ability to form abstract, high-level comprehension over extended conversations. For example, when a user narrates a long story, the agent should retain the content and derive a holistic understanding rather than just recall isolated facts.
We define Long-Range Understanding (LRU) as the ability to reason about long-form inputs and answer high-level questions that require an understanding of the overall content, rather than detailed recall. An example question might be: “Summarize the main experiences of Harry Potter.”

\subsubsection{Details on LRU datasets}

We evaluate LRU via the Summarization task \texttt{En.Sum} from $\infty$-Bench~\citep{infinitebench}. We follow the settings from ~\citep{HELMET} and use the GPT-4o model in evaluating the summarized text. In this process, we assess the fluency of the input text (scored as 0 or 1) and use the dot product of this score with the F1 score as the final evaluation metric.

\subsection{Selective Forgetting (SF)}

\subsubsection{Definition of SF} In long-term interactions, agents often face evolving or conflicting information—whether about the external world (e.g., changes in political leadership) or user-specific facts (e.g., a new occupation). This challenge is closely related to model editing \citep{memit, alphaedit} and knowledge unlearning \citep{Large_Scale_Knowledge_Washing}, which focus on modifying or removing factual knowledge from language models.
We define Selective Forgetting (SF) as the agent’s ability to detect and resolve contradictions between out of date knowledge and newly acquired information, ensuring the agent remains aligned with current realities and user states.
SF is distinct from Abstractive Retrieval (AR) in two key ways. (1) Certain questions requiring SF cannot be answered solely through AR. As illustrated in Figure~\ref{fig:intro}, an agent that retrieves all facts related to \texttt{pears} may fail to identify the updated information in the second message. (2) In AR, earlier messages remain relevant and should be retained, even when multiple pieces of evidence are required. In contrast, SF involves identifying outdated or incorrect information and discarding it. That is, AR requires preservation of all related content, whereas SF requires overwriting prior facts to reflect the most up-to-date truth.

\subsubsection{Details on SF datasets}

We use counterfactual edit pairs from the MQUAKE~\citep{mquake} dataset. Each sentence containing information was assigned a number. For each edit pair, the sentence representing outdated information (the distractor) is given a smaller number, while the sentence representing more recent information (the one containing the answer) is given a larger number. We then concatenate these sentences into a long context in order according to their assigned numbers. We evaluate the SF via two datasets: \textbf{Single-Hop FactConsolidation} and \textbf{Multi-Hop FactConsolidation}. In these tasks, the agent's responses are mostly informational entities. Therefore, we also use SubEM (Substring Exact Match) as the evaluation metric to calculate the accuracy of QA.

\subsection{Justification for competencies based on cognitive science}

Accurate retrieval is central to human memory research, as evidenced by classical forgetting curves and recall tests that foreground fidelity of recall~\citep{Ebbinghaus2013Memory}. However, a sole focus on accuracy obscures another fundamental axis: the timescale of learning and consolidation. Ebbinghaus observed that an initial, fleeting grasp rarely endures without reinforcement~\citep{Ebbinghaus2013Memory}, and ~\citet{James1890PrinciplesV1} distinguished primary (immediate) from secondary (enduring) memory. These classic distinctions ground our notions of test-time learning (incorporation of new information via memory) and long-range understanding (durable, integrated knowledge). Consistent with this, the Complementary Learning Systems (CLS) framework delineates a hippocampal system for rapid episodic learning and a neocortical system for gradual, structured knowledge accumulation, underscoring the need to assess both quick memorization and long-horizon retention~\citep{McClelland1995CLS}.

Beyond the acquisition–consolidation axis, another equally fundamental challenge is selective forgetting. Overlapping or contradictory traces can impede retrieval, and interference has long been recognized as a primary driver of forgetting in cognitive psychology~\citep{AndersonNeely1996Interference}. Neurocognitive evidence further shows that the brain engages targeted control mechanisms to resolve such interference at retrieval time~\citep{Wimber2015AdaptiveForgetting}. We therefore include selective forgetting—the ability to handle interference and contradictions—as a core dimension.

In sum, our four categories—accurate retrieval, test-time learning, long-range understanding, and selective forgetting—align with key dimensions of memory identified in cognitive science and AI memory systems, covering the essential capabilities that any robust memory mechanism must support in practice. Notably, the challenge of retaining previously acquired knowledge while incrementally accommodating new categories has also been extensively studied in continual learning and open-world discovery (\cite{fengprism}; \cite{feng2026generalized}), where models must balance stability and plasticity under distributional shift—a tension that directly parallels the interplay between accurate retrieval and selective forgetting in our framework. More broadly, learning from limited or weak supervision (\cite{he2024weakly}) and systematic surveys of emerging methodologies (\cite{he2024diffusion}) have reinforced the value of unified evaluation frameworks in driving progress across AI sub-fields, further motivating our benchmark design.

\begin{table}
\centering
\small           
\caption{Datasets categorized by the specific aspects of evaluation. Here 1K is 1024. }
\vspace{5pt}
\begin{tabularx}{0.92\linewidth}{@{}lXccc@{}}
\toprule
\textbf{Capability} & \textbf{Tasks} & \textbf{\# of Sequences : QAs}  & \textbf{Avg Len} \\
\midrule
\multirow{4}{*}{\parbox{3cm}{Accurate\\Retrieval}} 
    & SH-Doc QA           & 1 : 100 & 197K \\
    & MH-Doc QA           & 1 : 100 & 421K \\
    & LongMemEval (S*)   & 5 : 300 & 355K \\
    & EventQA           & 5 : 500 & 534K \\[2pt]
\midrule
\multirow{6}{*}{\parbox{3cm}{Test-Time\\Learning}}
    & BANKING-77                                 & 1 : 100 &  \\
    & CLINC-150                                  & 1 : 100 & \\
    & NLU                                        & 1 : 100 & 103K \\
    & TREC (Coarse)                              & 1 : 100 & \\
    & TREC (Fine)                                & 1 : 100 & \\
    & Movie-Rec Redial                           & 1 : 200 & 1.44M \\[2pt]
\midrule
\multirow{2}{*}{\parbox{3cm}{Long-Range \\Understanding}}
    & $\infty$Bench-Sum   & 100 : 100 & 172K \\ 
    & Detective QA   & 10 : 71 & 124K \\[2pt]
\midrule
\multirow{2}{*}{\parbox{3cm}{Selective Forgetting}}
    & \parbox{4cm}{FactConsolidation-SH}         & 1 : 100 & \multirow{2}{*}{262K} \\ 
    & \parbox{4cm}{FactConsolidation-MH}         & 1 : 100 & \\ 
\bottomrule
\end{tabularx}
\label{tab:datasets}
\end{table}

\section{Detailed Memory Agents Description}
\label{sec:agents_intro}

We give detailed description of the memory agents used in experiments in this section.

\subsection{Long-Context Agents}

We evaluate six modern long-context LLMs: GPT-4o~\citep{openai2025gpt4o} serves as the high-performance, low-latency model with better cost efficiency than prior generations. While GPT-4o-mini is a lightweight, budget-friendly alternative that enables large-scale evaluations by delivering faster responses and lower per-token costs. Notably, the GPT-4.1~\citep{openai2025gpt4.1mini} family strengthens instruction following and maintains strong performance at very large context windows (reported up to 1M tokens). Considering the higher token cost, we choose the GPT-4.1-mini in evaluation. GPT-5-mini is a compact reasoning variant of GPT-5\citep{openai2025gpt5mini}, offering a 400K-token context window with built-in chain-of-thought capabilities at reduced latency and cost. Gemini-2.0-Flash~\citep{deepmind2025gemini} targets high throughput and the use of built-in tools, offering a 1M token context window for efficient long-context processing. Claude-3.7-Sonnet~\citep{anthropic2025claude37} is a hybrid-reasoning model with optional visible “extended thinking,” strong math/coding skills, and developer-controlled thinking budgets. 

\subsection{RAG Agents}

We consider three RAG variants: \emph{Simple RAG Agents}, \emph{Embedding-based RAG Agents}, and \emph{Structure-Augmented RAG Agents}. 

\paragraph{ (1) Simple RAG Agents}  We implement a BM25~\citep{bm25} retriever as a strong lexical baseline: it scores documents by term frequency with saturation and inverse document frequency, with length normalization controlled by parameters $k_1$ and $b$. BM25 remains competitive for exact-match queries and complements dense retrievers with robust precision on keyworded questions.

\paragraph{ (2) Embedding-based RAG Agents} Contriever~\citep{contriever} is an unsupervised dense retriever trained via contrastive learning on large text corpora, enabling semantic matching without labeled pairs.  Text-Embedding-3-Small/Large~\citep{openai2024text} are OpenAI’s general-purpose embedding models offering a cost–quality trade-off (e.g., 1,536 vs. 3,072 dimensions) for search and retrieval. Qwen3-Embedding-4B~\citep{qwen3_embedding} is a 4B-parameter embedding/reranking model family geared toward multilingual retrieval and long-text understanding.

\paragraph{ (3) Structure-Augmented RAG Agents}  RAPTOR~\citep{sarthi2024raptor} is method building a hierarchical tree of recursive summaries (bottom-up clustering and abstraction) and retrieves across levels for long-document QA. GraphRAG~\citep{graphrag} extracts a knowledge graph and community hierarchy, then performs graph-aware retrieval and summarization. MemoRAG~\citep{memorag} introduces a dual-system pipeline with a light “global-memory” model to guide retrieval and a stronger model for final answers. HippoRAG-v2~\citep{hipporag-v2} extends hippocampal-inspired retrieval to improve factual, sense-making, and associative memory tasks over standard RAG. We also evaluate three open-sourced memory agents: Mem0, Cognee and Zep. Mem0~\citep{Mem0} provides a persistent agent memory layer for storing/retrieving user-specific knowledge to enhance personalization. Cognee~\citep{cognee} is an open-source memory engine that builds structured (graph-native) memories via ECL pipelines to power graph-aware RAG. Zep~\citep{Zep} is a temporal knowledge-graph memory platform for agents, designed to assemble and retrieve long-term conversational and business context. Beyond pairwise graph structures, hypergraph-based architectures that capture higher-order group-wise relationships over temporal sequences (\cite{tan2025h3m}; \cite{tan2025magnet}) represent a promising direction for enhancing structure-augmented memory agents.

\subsection{Agentic Memory Agents}

For Agentic Memory Agents, We evaluate the Self-RAG~\citep{self-rag}, MemGPT~\citep{MemGPT}, and MIRIX~\citep{mirix} on our benchmark. Self-RAG use LLMs as the agent to decide when/what to retrieve and to critique its own outputs. MemGPT operates the hierarchical memory management, paging relevant snippets between short-term and long-term stores and using event-driven interrupts to maintain coherence and evolvability over extended interactions. MIRIX adopts a multi-agent memory architecture with six specialized memory types (Core, Episodic, Semantic, Procedural, Resource, Knowledge Vault) and a coordinator that orchestrates updates/retrieval across agents. 

For comparability, we standardize prompts, tool access, and settings (like retrieval TopK and input chunk size) across above all systems.

\section{Prompts}
\label{sec:prompts}

We introduce the examples of prompt used memory construction and task execution in this section.

\subsection{Instructions for Memory Construction}
\label{sub:instructions_for_memory_construction}

\begin{figure}[h!]
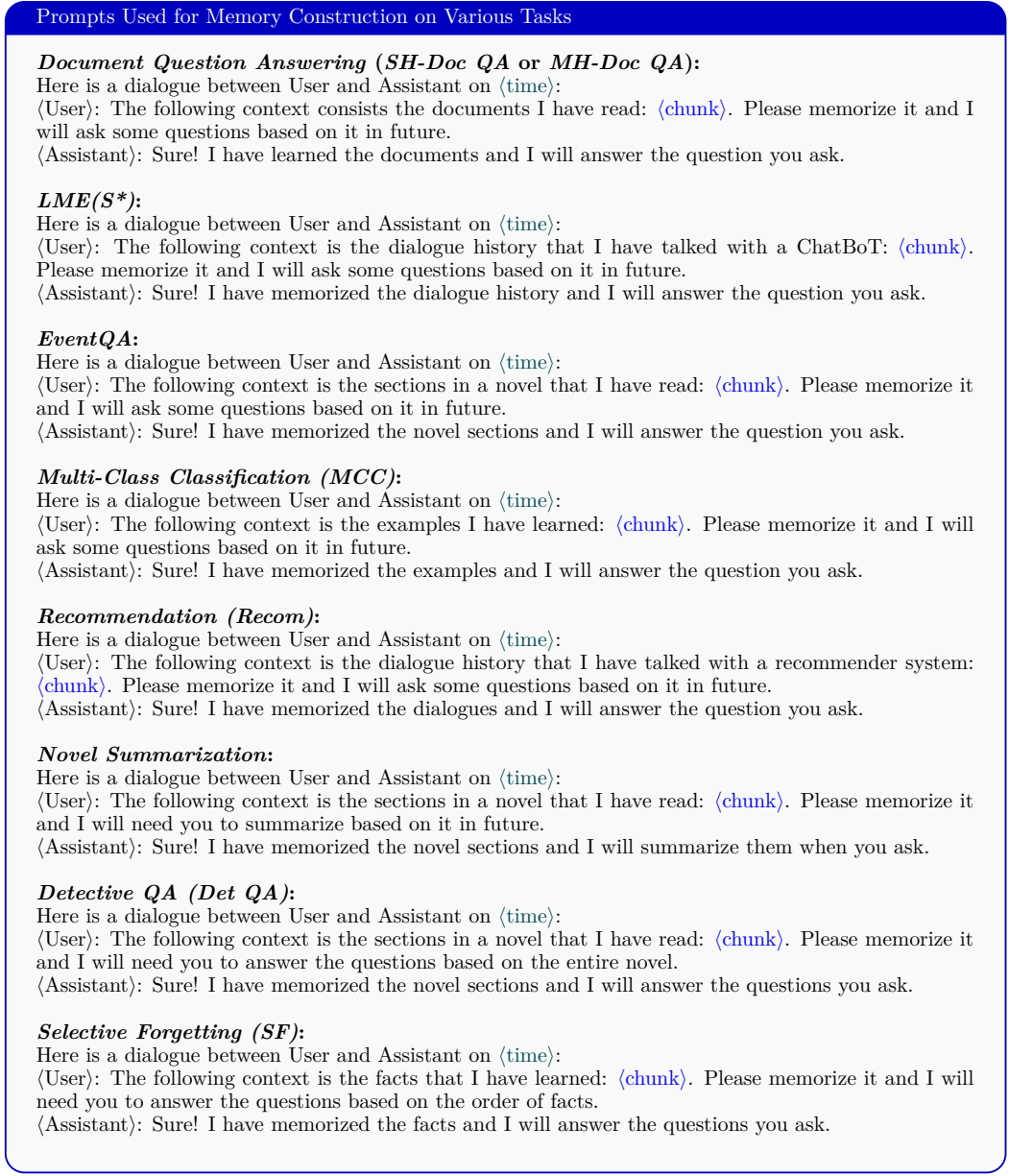

\centering
\resizebox{\textwidth}{!}{
\begin{tcolorbox}[colback=gray!5!white, colframe=blue!75!black, 
title=Prompts Used for Memory Construction on Various Tasks, boxrule=0.3mm, width=1.2\textwidth, arc=3mm, auto outer arc=true]
\textbf{\emph{Document Question Answering} (\emph{SH-Doc QA} or \emph{MH-Doc QA}):} \\
Here is a dialogue between User and Assistant on \textcolor{midnightgreen}{$\langle$time$\rangle$}:  \\
$\langle$User$\rangle$: The following context consists the documents I have read: \textcolor{blue}{$\langle$chunk$\rangle$}. Please memorize it and I will ask some questions based on it in future.   \\
$\langle$Assistant$\rangle$: Sure! I have learned the documents and I will answer the question you ask.  \\

\textbf{\emph{LME(S*)}:} \\
Here is a dialogue between User and Assistant on \textcolor{midnightgreen}{$\langle$time$\rangle$}:  \\
$\langle$User$\rangle$: The following context is the dialogue history that I have talked with a ChatBoT: \textcolor{blue}{$\langle$chunk$\rangle$}.  Please memorize it and I will ask some questions based on it in future.    \\
$\langle$Assistant$\rangle$:  Sure! I have memorized the  dialogue history and I will answer the question you ask.  \\

\textbf{\emph{EventQA}:} \\
Here is a dialogue between User and Assistant on \textcolor{midnightgreen}{$\langle$time$\rangle$}:  \\
$\langle$User$\rangle$: The following context is the sections in a novel that I have read: \textcolor{blue}{$\langle$chunk$\rangle$}.  Please memorize it and I will ask some questions based on it in future.    \\
$\langle$Assistant$\rangle$:  Sure! I have memorized the novel sections and I will answer the question you ask.  \\

\textbf{\emph{Multi-Class Classification (MCC)}:} \\
Here is a dialogue between User and Assistant on \textcolor{midnightgreen}{$\langle$time$\rangle$}:  \\
$\langle$User$\rangle$: The following context is the examples I have learned: \textcolor{blue}{$\langle$chunk$\rangle$}.  Please memorize it and I will ask some questions based on it in future.    \\
$\langle$Assistant$\rangle$:  Sure! I have memorized the examples and I will answer the question you ask.  \\

\textbf{\emph{Recommendation (Recom)}:} \\
Here is a dialogue between User and Assistant on \textcolor{midnightgreen}{$\langle$time$\rangle$}:  \\
$\langle$User$\rangle$: The following context is the dialogue history that I have talked with a recommender system: \textcolor{blue}{$\langle$chunk$\rangle$}.  Please memorize it and I will ask some questions based on it in future.    \\
$\langle$Assistant$\rangle$:  Sure! I have memorized the dialogues and I will answer the question you ask.  \\

\textbf{\emph{Novel Summarization}:} \\
Here is a dialogue between User and Assistant on \textcolor{midnightgreen}{$\langle$time$\rangle$}:  \\
$\langle$User$\rangle$: The following context is the sections in a novel that I have read: \textcolor{blue}{$\langle$chunk$\rangle$}.  Please memorize it and I will need you to summarize based on it in future.    \\
$\langle$Assistant$\rangle$:  Sure! I have memorized the novel sections and I will summarize them when you ask.  \\

\textbf{\emph{Detective QA (Det QA)}:} \\
Here is a dialogue between User and Assistant on \textcolor{midnightgreen}{$\langle$time$\rangle$}:  \\
$\langle$User$\rangle$: The following context is the sections in a novel that I have read: \textcolor{blue}{$\langle$chunk$\rangle$}.  Please memorize it and I will need you to answer the questions based on the entire novel.    \\
$\langle$Assistant$\rangle$:  Sure! I have memorized the novel sections and I will answer the questions you ask.  \\

\textbf{\emph{Selective Forgetting (SF)}:} \\
Here is a dialogue between User and Assistant on \textcolor{midnightgreen}{$\langle$time$\rangle$}:  \\
$\langle$User$\rangle$: The following context is the facts that I have learned: \textcolor{blue}{$\langle$chunk$\rangle$}.  Please memorize it and I will need you to answer the questions based on the order of facts.    \\
$\langle$Assistant$\rangle$:  Sure! I have memorized the facts and I will answer the questions you ask.  \\

\end{tcolorbox}}
\caption{The prompts we use for the agents to create the memory.}
\label{fig:prompt_memory_construction}
\end{figure}

When processing long-context inputs, we split the content into chunks of a specified size and feed these chunks into the agent as memory. The agent can then extract relevant information from its memory based on the query to assist with query execution. This chunking approach helps organize and manage large amounts of contextual information, making retrieval and reasoning more efficient. In Figure ~\ref{fig:prompt_memory_construction}, we provide several example instructions that require the agent to memorize the corresponding context.

\subsection{Instructions for Task Execution}

\begin{figure}[h!]
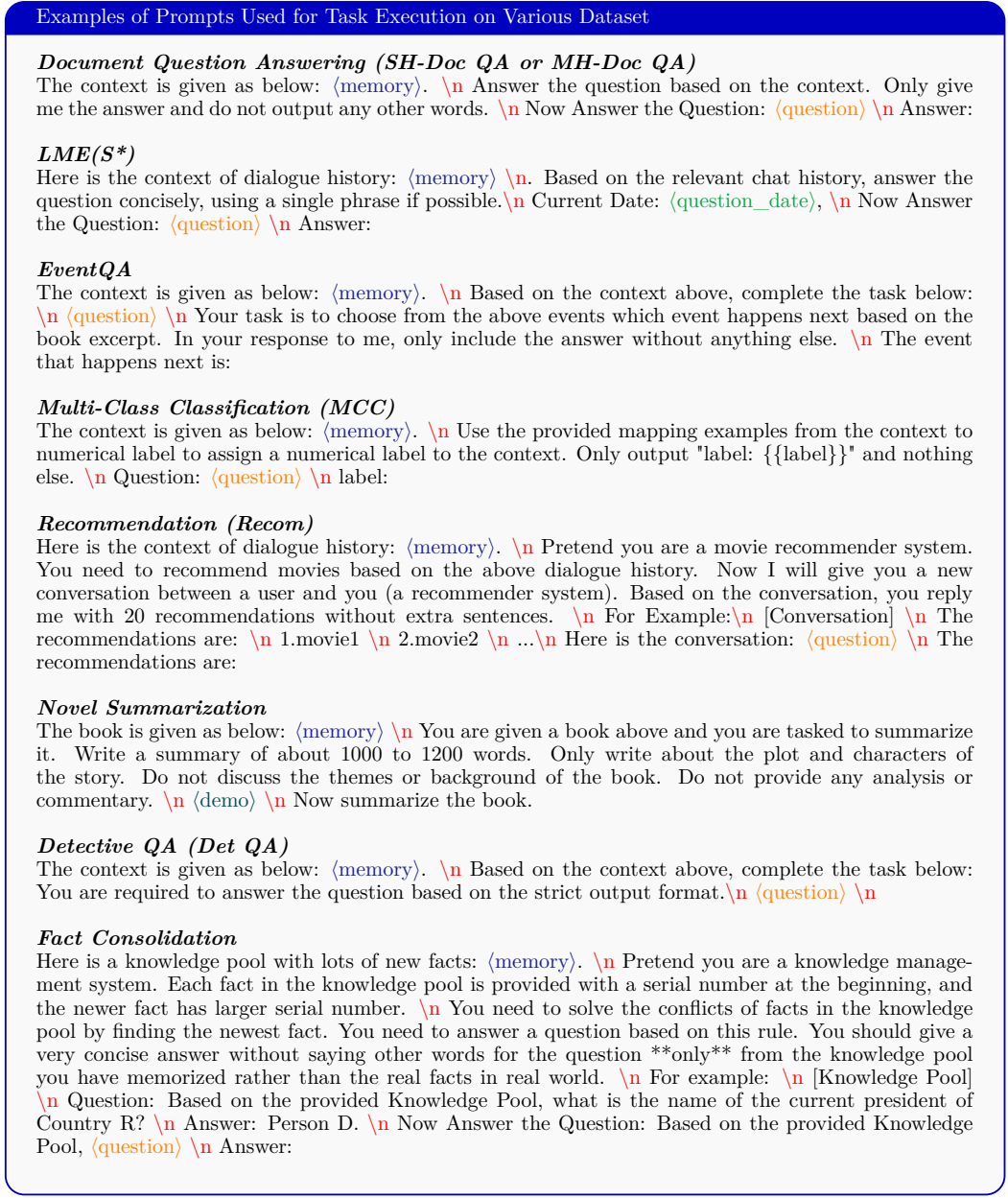

\centering
\resizebox{\textwidth}{!}{
\begin{tcolorbox}[colback=gray!5!white, colframe=blue!75!black, 
title=Examples of Prompts Used for Task Execution on Various Dataset, boxrule=0.3mm, width=1.2\textwidth, arc=3mm, auto outer arc=true]

\emph{\textbf{Document Question Answering (SH-Doc QA or MH-Doc QA)}} \\ 
The context is given as below: \textcolor{deepblue}{$\langle$memory$\rangle$}. \textcolor{red}{\textbackslash n} Answer the question based on the context. Only give me the answer and do not output any other words. \textcolor{red}{\textbackslash n} Now Answer the Question: \textcolor{orange}{$\langle$question$\rangle$} \textcolor{red}{\textbackslash n} Answer:\\

\emph{\textbf{LME(S*)}} \\ 
Here is the context of dialogue history: \textcolor{deepblue}{$\langle$memory$\rangle$}  \textcolor{red}{\textbackslash n}. Based on the relevant chat history, answer the question concisely, using a single phrase if possible.\textcolor{red}{\textbackslash n} Current Date: \textcolor{deepgreen}{$\langle$question\_date$\rangle$}, \textcolor{red}{\textbackslash n} Now Answer the Question: \textcolor{orange}{$\langle$question$\rangle$} \textcolor{red}{\textbackslash n} Answer: \\ 

\emph{\textbf{EventQA}} \\ 
The context is given as below: \textcolor{deepblue}{$\langle$memory$\rangle$}. \textcolor{red}{\textbackslash n} Based on the context above, complete the task below: \textcolor{red}{\textbackslash n} \textcolor{orange}{$\langle$question$\rangle$} \textcolor{red}{\textbackslash n} Your task is to choose from the above events which event happens next based on the book excerpt. In your response to me, only include the answer without anything else. \textcolor{red}{\textbackslash n} The event that happens next is: \\

\emph{\textbf{Multi-Class Classification (MCC)}} \\ 
The context is given as below: \textcolor{deepblue}{$\langle$memory$\rangle$}. \textcolor{red}{\textbackslash n} Use the provided mapping examples from the context to numerical label to assign a numerical label to the context. Only output "label: \{\{label\}\}" and nothing else. \textcolor{red}{\textbackslash n} Question:  \textcolor{orange}{$\langle$question$\rangle$} \textcolor{red}{\textbackslash n} label: \\

\emph{\textbf{Recommendation (Recom)}} \\ 
Here is the context of dialogue history: \textcolor{deepblue}{$\langle$memory$\rangle$}. \textcolor{red}{\textbackslash n} Pretend you are a movie recommender system. You need to recommend movies based on the above dialogue history. Now I will give you a new conversation between a user and you (a recommender system). Based on the conversation, you reply me with 20 recommendations without extra sentences. \textcolor{red}{\textbackslash n} For Example:\textcolor{red}{\textbackslash n} [Conversation] \textcolor{red}{\textbackslash n} The recommendations are: \textcolor{red}{\textbackslash n} 1.movie1 \textcolor{red}{\textbackslash n} 2.movie2 \textcolor{red}{\textbackslash n} ...\textcolor{red}{\textbackslash n} Here is the conversation: \textcolor{orange}{$\langle$question$\rangle$} \textcolor{red}{\textbackslash n} The recommendations are: \\

\emph{\textbf{Novel Summarization}} \\ 
The book is given as below: \textcolor{deepblue}{$\langle$memory$\rangle$} \textcolor{red}{\textbackslash n} You are given a book above and you are tasked to summarize it. Write a summary of about 1000 to 1200 words. Only write about the plot and characters of the story. Do not discuss the themes or background of the book. Do not provide any analysis or commentary. \textcolor{red}{\textbackslash n} \textcolor{midnightgreen}{$\langle$demo$\rangle$} \textcolor{red}{\textbackslash n} Now summarize the book.  \\

\emph{\textbf{Detective QA (Det QA)}} \\ 
The context is given as below: \textcolor{deepblue}{$\langle$memory$\rangle$}. \textcolor{red}{\textbackslash n}  Based on the context above, complete the task below: You are required to answer the question based on the strict output format.\textcolor{red}{\textbackslash n}  \textcolor{orange}{$\langle$question$\rangle$} \textcolor{red}{\textbackslash n}   \\

\emph{\textbf{Fact Consolidation}} \\ 
Here is a knowledge pool with lots of new facts: \textcolor{deepblue}{$\langle$memory$\rangle$}. \textcolor{red}{\textbackslash n} Pretend you are a knowledge management system. Each fact in the knowledge pool is provided with a serial number at the beginning, and the newer fact has larger serial number. \textcolor{red}{\textbackslash n} You need to solve the conflicts of facts in the knowledge pool by finding the newest fact. You need to answer a question based on this rule. You should give a very concise answer without saying other words for the question **only** from the knowledge pool you have memorized rather than the real facts in real world. \textcolor{red}{\textbackslash n} For example: \textcolor{red}{\textbackslash n} [Knowledge Pool] \textcolor{red}{\textbackslash n} Question: Based on the provided Knowledge Pool, what is the name of the current president of Country R? \textcolor{red}{\textbackslash n} Answer: Person D. \textcolor{red}{\textbackslash n} Now Answer the Question: Based on the provided Knowledge Pool, \textcolor{orange}{$\langle$question$\rangle$} \textcolor{red}{\textbackslash n} Answer:    \\

\end{tcolorbox}}
\caption{The example prompts we use for the \emph{Memory Agents} in Table \ref{tab:overall_performance_comparison}. Here $\langle$memory$\rangle$ refers to the accumulated text from the sequential inputs.}
\label{fig:prompt_task_execution_agent}
\end{figure}

In Figure ~\ref{fig:prompt_task_execution_agent}, we provide the examples of instructions used on different of datasets when handling the input queries. For some existing datasets, we refer the prompt settings from previous work such as ~\citep{RULER, longmemeval}. For the dataset \textbf{$\infty$Bench-Sum}, we also insert two answer examples as \textcolor{midnightgreen}{$\langle$demo$\rangle$} in the prompt to help the agent better understand the questions and standardize its outputs.

\section{Detailed Experimental Results}
\label{sec:detailed_exp_results}

In this section, we provide detailed versions of the results presented in the main text.

\begin{table*}[t]
    \centering
    \small
     \caption{Overall performance comparison on the datasets for TTL. All RAG agents and commercial memory agents use GPT-4o-mini as the backbone. }
    \resizebox{0.66\textwidth}{!}{
    \begin{tabular}{l|ccccc}
        \rowcolor{headergray}
        \rowcolor{headergray}
        \textbf{Agent Type} & BANKING & CLINIC & NLU & TREC C & TREC F \\
        \toprule
        \rowcolor{groupblue}
        \multicolumn{6}{c}{\emph{Long-Context Agents}} \\
        GPT-4o            & 96.0 & 96.0 & \textbf{90.0} & 87.0 & 69.0   \\
        GPT-4o-mini       & 93.0 & 93.0 & 87.0 & 73.0 & 66.0   \\
        GPT-4.1-mini      & 93.0 & 82.0 & 85.0 & 68.0 & 50.0  \\
        GPT-5-mini      & 88.0 & 92.0 & 88.0 & \textbf{88.0} & 64.0  \\
        Gemini-2.0-Flash  & 91.0 & 90.0 & 84.0 & \textbf{88.0} & 67.0  \\
        Claude-3.7-Sonnet & \textbf{97.0} & \textbf{98.0} & 86.0 & 87.0 & \textbf{79.0} \\
    
        \midrule
        \midrule
        \rowcolor{headergray}
        GPT-4o-mini & 93.0 & 93.0 & 87.0 & 73.0 & 66.0  \\
        \rowcolor{groupgreen}
        \multicolumn{6}{c}{\emph{Simple RAG Agents}} \\
        BM25              & 89.0 & 89.0 & 84.0 & 62.0 & 53.0  \\
        \rowcolor{groupgreen}
        \multicolumn{6}{c}{\emph{Embedding RAG Agents}} \\
        Contriever         & 89.0 & 88.0 & 80.0 & 55.0 & 41.0  \\
        Text-Embed-3-Small & 88.0 & 89.0 & 83.0 & 54.0 & 36.0  \\
        Text-Embed-3-Large & \textbf{90.0} & \textbf{91.0} & 80.0 & 55.0 & 46.0  \\
        Qwen3-Embedding-4B   & \textbf{90.0}  & 88.0  & \textbf{86.0}  & \textbf{67.0}  & \textbf{59.0}     \\
        \rowcolor{groupgreen}
        \multicolumn{6}{c}{\emph{Structure-Augmented RAG Agents}} \\
        RAPTOR      & 78.0 & 75.0 & 73.0 & 48.0 & 23.0  \\
        GraphRAG    & 64.0 & 54.0 & 49.0 & 24.0 & 6.0    \\
        MemoRAG     & 90.0 & 87.0 & 86.0 & 66.0 & 56.0   \\
        HippoRAG-v2 & 81.0 & 86.0 & 73.0 & 38.0 & 29.0   \\
        Mem0        & 35.0 & 37.0 & 35.0 & 29.0 & 26.0        \\
        Cognee      & 34.0 & 42.0 & 42.0 & 41.0 & 18.0   \\
        Zep         & 83.0 & 74.0 & 70.0 & 50.0 & 37.0   \\
        \rowcolor{grouppink}
        \multicolumn{6}{c}{\emph{Agentic Memory Agents}}    \\
        Self-RAG       & 19.0 & 13.0 & 6.0 & 15.0 & 5.0      \\
        MemGPT         & 89.0 & 83.0 & 79.0 & 56.0 & 31.0    \\
        MIRIX                & 42.0 & 53.0 & 49.0 & 36.0 & 12.0   \\
        MIRIX(4.1-mini)      & 65.0 & 83.0 & 69.0 & 73.0 & 25.0    \\  
        \bottomrule
    \end{tabular}}
\label{tab:TTL_overall_performance_comparison}
\end{table*}

\subsection{Detailed Results on TTL}

We give detailed results on Multi-Class Classification (MCC) task in Table ~\ref{tab:TTL_overall_performance_comparison}. For all three types of tasks, RAG-based agents generally underperform compared to their respective GPT-4o-mini backbones. This observation highlights certain limitations inherent to the RAG approach. For instance, in TTL tasks, RAG-based methods often struggle to more accurately retrieve context from memory that is closely associated with the input.

\subsection{Results on Input Chunk Size Ablation Study}
\label{sub:input_chunk_ablation}

In Table ~\ref{tab:different_chunksize_full}, we report the detailed results on evaluating the RAG-based Agents on different chunk sizes and datasets. We selected chunk sizes from the two sets \{512, 4096\} and \{512, 1024, 2048, 4096\}. 

\subsection{Results on Retrieval TopK Ablation Study}
\label{sub:results_on_topk}

In Table ~\ref{tab:retrieval_topk_ablation}, we report the detailed results of the selected RAG-based Agents evaluated on five datasets. We choose different TopK ranging from \{2, 5, 10\}.

\subsection{Results on Different Context Length Ablation Study}
\label{sub:different_context_length}

In Table ~\ref{tab:different_context_length_full}, we report the performances of different agents when scaling the input length. We measure the average context length via the tokenizer of GPT-4o-mini and here 1K is 1024. For Long-Context Agents, tasks in the AR series generally achieve satisfactory performance at relatively small context lengths (e.g., around 50K tokens). However, as the context length increases, the performance of these agents declines accordingly. In contrast, for the RAG-based agents Mem0 and Cognee, their performance is significantly lower than that of their backbone, GPT-4o-mini, even when the context length is relatively small.

\begin{table*}[t]
    \centering
    \caption{Performance comparison on different datasets and chunk sizes. Here we choose chunk sizes from \{512, 1024, 2048, 4096\} and we use k=10 for RAG-based methods.}
    \resizebox{\textwidth}{!}{
    \begin{tabular}{c|cccc|cccc|cccc}
        \toprule
          & \multicolumn{4}{c|}{SH-Doc QA} & \multicolumn{4}{c|}{MH-Doc QA} & \multicolumn{4}{c}{$\infty$Bench-Sum}  \\
           & 512 & 1024 & 2048 & 4096 & 512 & 1024 & 2048 & 4096 & 512 & 1024 & 2048 & 4096  \\
          \midrule
          BM25              & 66.0 & 67.0 & 68.0 & 66.0 & 56.0 & 54.0 & 52.0 & 56.0 & 11.5 & 13.2 & 15.2 & \textbf{19.0} \\
          Qwen3-Embedding-4B  & 57.0 & 53.0 & 52.0 & 50.0 & 47.0 & 44.0 & 40.0 & 38.0 & 7.9 & 9.4 & 13.2 & 14.8  \\

          HippoRAG-v2       & 76.0 & 70.0 & 57.0 & 49.0 & 66.0 & 63.0 & 51.0 & 38.0 & 4.6 & 6.0 & 10.5 & 14.6 \\
          MemGPT            & 41.0 & 32.0 & 24.0 & 27.0 & 38.0 & 33.0 & 37.0 & 35.0 & 1.2 & 1.8 & 4.2 & 2.5   \\
          \bottomrule
        \end{tabular}}
    \label{tab:different_chunksize_full}
     \centering
    \vspace{2pt}
    \caption{Performance comparison on different retrieve number.}
    \resizebox{0.99\linewidth}{!}{
    \begin{tabular}{c|ccc|ccc|ccc|ccc}
        \toprule
          & \multicolumn{3}{c|}{SH-Doc QA} & \multicolumn{3}{c|}{MH-Doc QA}  & \multicolumn{3}{c|}{EventQA}   & \multicolumn{3}{c}{TTL (MCC)}   \\
          & R=2 & R=5 & R=10  & R=2 & R=5 & R=10 & R=2 & R=5 & R=10 & R=2 & R=5 & R=10    \\

          \midrule
          BM25              & 50.0 & 60.0 & 66.0 &  49.0 & 54.0 &  56.0 & 66.6 & 71.2 & 74.6 & 67.8 & 74.6 & 75.4 \\
          Contriever        &  17.0 & 20.0 & 22.0 & 22.0 & 27.0 & 31.0 & 54.4 & 66.8 & 56.0 & 63.0 & 70.0 & 70.6 \\
          
          Text-Embed-3-Large & 36.0 & 47.0 & 54.0 & 37.0 & 41.0 & 44.0 & 51.8 & 62.4 & 70.0 & 59.4 & 69.4 & 72.4 \\

          RAPTOR              & 22.0 & 27.0 & 29.0 & 30.0 & 36.0 & 38.0  & 45.8 & 41.8 & 40.4 & 56.3 & 57.4 & 59.4 \\
          HippoRAG-v2         & 60.0 & 69.0 & 76.0 & 53.0 & 60.0 & 66.0 & 58.8 & 67.6 & 67.4 & 58.8 & 61.4 & 61.4 \\
          Self-RAG           & 27.0 & 33.0 & 35.0 & 34.0 & 39.0 & 42.0  & 28.2 & 30.6 & 31.8  & 9.0 & 11.6 & 11.6 \\
        
          \bottomrule
        \end{tabular}
        }
    \label{tab:retrieval_topk_ablation}
    \centering
    \vspace{5pt}
    \caption{Performance comparison on different context length.}
    \resizebox{0.98\linewidth}{!}{
    \begin{tabular}{c|ccc|ccc|ccc|ccc|ccc}
        \toprule
          & \multicolumn{3}{c|}{SH-Doc QA} & \multicolumn{3}{c|}{MH-Doc QA} & \multicolumn{3}{c|}{EventQA}   & \multicolumn{3}{c|}{FactCon-SH} & \multicolumn{3}{c}{FactCon-MH} \\
          & 51K & 104K & 197K & 51K & 104K & 421K & 51K & 108K & 534K & 32K & 64K  & 262K & 32K & 64K  & 262K \\
          \midrule
          GPT-4o                & 91.0  & 84.0 & 72.0 & 72.0  & 68.0  & 51.0 & 96.8 & 94.0 & 77.2 & 88.0 & 85.0 & 60.0 & 10.0 & 13.0 & 5.0 \\
          
          GPT-4o-mini           & 84.0 & 83.0 & 64.0 & 58.0 & 54.0 & 43.0 & 90.2 & 85.8 & 59.0 & 63.0 & 58.0 & 45.0 & 10.0 & 5.0 & 5.0 \\
          
          GPT-4.1-mini          & 93.0  & 86.0  & 83.0  & 72.0 & 75.0 & 66.0 & 97.0 & 93.8 & 82.6 & 82.0 & 72.0 & 36.0 & 7.0 & 9.0 & 5.0 \\
          
          Gemini-2.0-Flash      & 92.0  & 87.0  & 87.0  & 69.0 & 61.0 & 59.0 & 93.4 & 88.6 & 67.2 & 49.0 & 62.0 & 30.0 & 7.0 & 9.0 & 3.0 \\
          
          Claude-3.7-Sonnet     & 90.0 & 82.0 & 77.0 & 67.0 & 59.0 & 53.0 & 96.6 & 95.2 & 74.6 & 46.0 & 45.0 & 43.0 & 2.0 & 2.0 & 2.0  \\
          
          \midrule
          Mem0       & 31.0  & 25.0  & 25.0  & 36.0 & 29.0  & 32.0   & 60.8 & 47.0 & 37.5 & 22.0 & 8.0 & 18.0 & 3.0 & 2.0 & 2.0  \\
          Cognee     & 38.0  & 42.0  & 31.0 &  36.0 & 38.0 &  26.0  & 53.4 & 39.0 & 26.8 & 39.0 & 31.0 & 28.0 & 4.0 & 5.0 & 3.0  \\
          \bottomrule
        \end{tabular}
        }
    \label{tab:different_context_length_full}
     
\end{table*}

\subsection{Results on Computational Latency Analysis}
\label{sub:latency_analysis}

To illustrate the latency of various memory agents in terms of (1) Memory Construction (M.C.); (2) Query Execution (Q.E.), we report the latency of various memory agents on MH-QA and LME (S*). This part of experiments is done on a server with Four NVDIA L40 GPU and AMD EPYC 7713 64-Core CPU. 
We use the NV-Embed-v2 (7B) as the embedding model in HippoRAG-v2. We show the results in Table \ref{tab:computation_latency_longcontext} and \ref{tab:computation_latency_rag}. From the table, we find that using a smaller chunk size requires significantly more time for memory construction, especially for methods such as HippoRAG-v2, Mem0, Cognee, and MemGPT. Meanwhile, methods such as Mem0, Cognee and MIRIX need extremely high resources when constructing the memory.

\begin{table*}[t]
    \centering
    \caption{Computational latency (in seconds) comparison on Long-Context Agents.}
    \resizebox{0.50\linewidth}{!}{
    \begin{tabular}{c|c|c}
        \toprule
          & MH-QA & LME (S*)  \\
          
          \midrule
          GPT-4o               & 17.0 & 20.1  \\
          GPT-4o-mini          & 4.9  & 5.4       \\
          GPT-4.1-mini         & 9.0  & 7.4       \\
          Gemini-2.0-Flash     & 12.4 & 10.1       \\
          Claude-3.7-Sonnet    & 23.3 & 22.7       \\
          
          \bottomrule
        \end{tabular}
        }
    \label{tab:computation_latency_longcontext}
    \vspace{10pt}
    \centering
    \caption{Computational latency (in seconds) comparison on RAG based agents. M.C. means Memory Construction and Q.E. means Query Execution. *Indicates that the time is obtained through estimation. }
    \resizebox{0.80\linewidth}{!}{
    \begin{tabular}{c|cccc|cccc}
        \toprule
          & \multicolumn{4}{c|}{MH-QA} & \multicolumn{4}{c}{LME (S*)}  \\

          & \multicolumn{2}{c}{512} & \multicolumn{2}{c|}{4096}  & \multicolumn{2}{c}{512} & \multicolumn{2}{c}{4096} \\
          
          \textbf{} & M.C. & Q.E. & M.C. & Q.E. & M.C. & Q.E. & M.C. & Q.E.   \\
          
          \midrule
          BM25                & 0.12 & 0.47 & 0.11 & 1.7 & 0.09 & 1.1 & 0.08 & 1.9 \\
          
          \midrule
          Contriever          & 7.4 & 0.59 & 1.7 & 2.0 & 6.9 & 0.92 & 1.6 & 1.9 \\
          Text-Embed-3-Large  & 6.1 & 0.46 & 5.0 & 1.7 & 6.5 & 0.62 & 5.8 & 1.8 \\
          Qwen3-Embedding-4B   & 367  & 0.49 & 470 & 1.9 & 293 & 0.71 & 372 & 1.8           \\

          \midrule
          RAPTOR              & 193 & 0.41 & 161 & 0.67 & 108 & 0.60 & 104 & 0.53     \\
          GraphRAG            & 97.8 & 12.8 & 91.9 & 10.9 & 149 & 7.0 & 88.8 & 7.8    \\
          HippoRAG-v2         & 1089  & 0.71 & 380  & 1.71 & 544 & 1.5  & 188 & 3.5         \\
          Mem0                & 10804 & 0.79 & 1334 & 0.65 & 18483 & 1.6 & 2946 & 1.7 \\
          Cognee              & 11890 & 58.7 & 1185 & 4.8 & 4728 & 7.7 & 738 & 4.1     \\

          \midrule
          Self-RAG            & 11.4 & 3.1 & 8.1 & 2.4 & 5.3 & 0.82 & 5.2 & 1.0    \\
          MemGPT              & 433 & 9.4 & 101 & 10.5 & 392 & 11.7 & 85.5 & 12.3  \\
          MIRIX    & 29000* & - & 20171 & 14.1 & 12600* & - & 3258  & 8.7  \\
          MIRIX (GPT-4.1-mini) & 28800* & - & 21361 & 16.9 & 9000* & - & 2512 & 9.2 \\ 
          \bottomrule
        \end{tabular}
        }
    \label{tab:computation_latency_rag}
    \centering
\end{table*}

\subsection{GPU Memory Usage Comparison}
\label{sub:gpu_memory_usage_comparison}
In main experiments, we mostly use the LLM API as the backbone models which do not need local GPUs. In our experiments, the HippoRAG-v2 (NV-Embed-v2) and Qwen3-Embedding-4B require running the embedding model on GPU. We report their peak GPU memory usage in Table ~\ref{tab:gpu_memory_report}, where all experiments are conducted on a single A100 80GB GPU.

\begin{table*}
    \centering
    \caption{Peak GPU memory usage of embedding models (MB). We measure the memory usage on MH-QA dataset with different chunk size.}
    \resizebox{0.6\linewidth}{!}{
    \begin{tabular}{c|c|c}
        \toprule
         Agents / Chunk Size & 512 & 4096 \\
        \midrule
        HippoRAG-v2 (NV-Embed-v2) & 27674  & 60205  \\
        Qwen3-Embedding-4B & 16671 & 41262     \\    \bottomrule
    \end{tabular}
    }
\label{tab:gpu_memory_report}
\end{table*}

\section{Experimental Settings}
\label{sec:exp_settings}

In this section, we present the experimental settings in evaluation.

\subsection{Max Output Tokens}

We provide the token number limitation for each task in Table ~\ref{tab:token_limits}.

\subsection{Settings of the RAG Agents}

For the embedding model selection in Structure-Augmented RAG Agents and Agentic Memory Agents, most approaches utilize OpenAI's embedding models, such as Text-Embed-3-Small. While for the HippoRAG-v2 method, we follow the same experimental setting as in \citet{hipporag-v2}, employing the NV-Embed-v2 model.  

We implement three open-sourced memory agents in our main experiments. (1) For Mem0, we use \textbf{memory.add()} function to add the message with the content from each context chunk into the agent's memory repository during memory consolidation. During query execution, the relevant memory elements are retrieved through \textbf{memory.search()} function. The retrieved memories are then integrated into the query before being processed by the GPT-4o-mini backbone model to complete the requested tasks. (2) For MemGPT, we employ the \textbf{insert\_passage()} function during the memory consolidation phase to inject long context chunks into the Archival Memory structure. During query execution, this agent processes requests via the \textbf{send\_message()} function which generates appropriate responses based on the archived information. (3) For Cognee, we utilize the \textbf{cognee.add()} and \textbf{cognee.cognify()} functions to construct the memory graph from input chunks wherein the memory consolidation phase. During query execution, the \textbf{cognee.search()} function is used to retrieve contextually relevant information from the memory graph based on the input query.

\subsection{Settings of the Chunk Size}

We use smaller chunk size (512) for synthetic context used in AR and SF. For some tasks based on continuous text, such as $\infty$Bench and EventQA, we used a larger chunk size (4096). For tasks such as MCC and Recom, considering the characteristics of these tasks and the computational cost, we also chose a larger chunk size (4096). For the memory construction methods that are more time-consuming and requiring more API cost, Mem0, Zep, Cognee and MIRIX, we uniformly used a chunk size of 4096 across all datasets. The detailed settings are presented in Table ~\ref{tab:different_context_length_setting}.

\begin{table*}
    \centering
    \caption{Maximum output token limits for various tasks}
    \resizebox{0.50\linewidth}{!}{
        \begin{tabular}{c|c}
        \toprule
            Task & Max Output Tokens \\
            \midrule
            SH-QA / MH-QA & 50 \\
            LME(S*) & 100 \\
            EventQA & 40 \\
            \midrule
            MCC & 20 \\
            Movie Recommendation & 300 \\
            \midrule
            $\infty$ Bench-Sum & 1,200 \\
            Detective QA & 500  \\
            \midrule
            FactConsolidation & 10 \\
          \bottomrule
        \end{tabular}
    }
    \label{tab:token_limits}
    \centering
    \vspace{10pt}
    \caption{The choice of chunk size for different datasets.} 
    \begin{tabular}{c|cc}
        \toprule
          \textbf{Chunk Size}  & 512 & 4096 \\
          \midrule
          & SH-QA, MH-QA  & $\infty$Bench-Sum \\
          Dataset      & FactCon-SH, FactCon-MH & MCC, Recom \\  
          & LME(S*) & EventQA, Detective QA   \\
          \bottomrule
        \end{tabular}

    \label{tab:different_context_length_setting}
\end{table*}

\section{Task Rationale and Justification for Selective Forgetting Task}
\label{app:selective_forgetting}

While the Selective Forgetting task may appear specialized or even synthetic at first glance, it is designed to address a fundamental, universal challenge in long-term memory systems: maintaining context efficiency and mitigating interference between outdated and updated information. We justify the design, novelty, and validity of this task along four tightly linked dimensions:
\begin{enumerate}
    \item \textbf{Theoretical Necessity}: For any memory system—biological or artificial—storage capacity is inherently finite. The ability to autonomously discard (or selectively forget) outdated, redundant, or superseded information is not a niche requirement, but a core prerequisite for keeping memory representations concise, robust, and free from conflicting signals. Our task is designed as a controlled proxy to evaluate this underexplored yet critical capability.
    
    \item \textbf{Distinction from Previous Settings (e.g., Knowledge Updating)}: While prior work has explored knowledge updating (which focuses on overwriting old facts with new, conflicting ones), our work uniquely emphasizes the explicit, proactive removal of non-essential information to free up cognitive and contextual space. This distinguishes our task from existing fact-updating benchmarks, and we view this framework as a foundational step toward evaluating more complex memory management behaviors in real-world settings.
    
    \item \textbf{Justification for the Controlled Synthetic Setting}: We acknowledge that the current task setup includes synthetic elements, and this design choice is deliberate and methodologically justified. Constructing a naturalistic dataset with long-term interaction history (over 100K tokens) and unambiguous, precise ground-truth annotations for what information should be forgotten is inherently challenging, as real-world forgetting decisions are often ambiguous and context-dependent. Our controlled synthetic setting isolates the selective forgetting capability from confounding factors, enabling reproducible, apples-to-apples comparisons across different memory agent architectures.
    
    \item \textbf{Validity and Feasibility of the Task}: Critically, we confirm that our proposed task, despite its synthetic design, is fully valid and solvable. As demonstrated in our ablation study, long-context agents with strong reasoning capabilities achieve near-perfect performance on short-context (6K tokens) versions of the dataset. This result directly confirms that performance degradation on long-context, full-length task inputs stems not from flaws in the task definition itself, but from the fundamental limitations of current memory agents in long-range reasoning—specifically, their inability to accurately identify and discard outdated information across extended interaction histories. Emerging approaches such as execution-grounded reinforcement learning for LLMs (\cite{li2026boostapr}) may offer a path toward strengthening such long-range reasoning capabilities in future memory agent designs.

\end{enumerate}

\section{Supplementary Validation for Test-Time Learning}
\label{app:test_time_learning}

This section provides supplementary justification for the terminology and design of our Test-Time Learning (TTL) evaluation paradigm, as well as additional zero-shot baseline experiments that validate the core premise of the TTL task.

\subsection{Terminology and Design Rationale}
We use the term \textit{Test-Time Learning (TTL)} to describe the class of tasks that evaluate an agent's ability to acquire task-specific skills and rules incrementally from interaction history, and apply these learned patterns to unseen inputs at inference time. The terminology and task design are justified by three core principles:

\begin{enumerate}
    \item \textbf{Distinguishing Skill Acquisition from Static Information Retrieval}: We explicitly differentiate TTL from standard retrieval-focused tasks (e.g., Accurate Retrieval, Long-Range Understanding) to highlight its unique focus on learning, rather than fact recall. In retrieval tasks, the agent's core goal is to fetch pre-defined, static facts from the interaction history. In contrast, TTL tasks (e.g., multi-class classification on Banking77, personalized movie recommendation) require the agent to induce latent classification rules, preference patterns, and task schemas from sequential labeled examples in the dialogue history, then generalize these patterns to completely novel, out-of-sample inputs. This operationalization directly aligns with the canonical definition of learning: updating a behavioral policy based on accumulated experience, which occurs exclusively during test-time interaction with no pre-training or fine-tuning on the target task. This notion of learning through interaction also connects to broader efforts in enhancing LLM capabilities via multi-agent collaborative reasoning (\cite{li2026draftrl}), where models must similarly extract generalizable patterns from limited or indirect signals rather than relying on pre-defined knowledge.

    \item \textbf{Controlled Operationalization of Online Learning}: While a fully dynamic online learning setting would involve an interleaved loop of memory update, retrieval, task execution, and feedback-driven memory refinement, such a setup introduces significant confounding variables that make it impossible to isolate an agent's memory capacity from unrelated reasoning, planning, or execution errors. To enable robust, reproducible, and unbiased evaluation of pure memory-enabled learning capability, we adopt a two-stage protocol that retains the core of online learning while eliminating confounds:
    \begin{itemize}
        \item \textbf{Acquisition Phase}: The agent incrementally processes a sequence of labeled task examples, simulating the experience accumulation process of real-world agent interactions over time.
        \item \textbf{Evaluation Phase}: The agent is tested on its ability to apply the rules and patterns acquired in the acquisition phase to held-out, unseen inputs.
    \end{itemize}
    This design ensures that performance differences between models can be directly attributed to their ability to leverage long-term interaction history for learning, rather than spurious factors from dynamic feedback loops.
    \item \textbf{Foundational Framework for Agent Memory Research}: As research on memory-enabled LLM agents remains in a nascent stage, our TTL evaluation framework provides a standardized, reproducible operationalization of in-situ learning for memory agents. While our current protocol simplifies the fully online learning loop for evaluation stability, it successfully captures the core competence required for real-world self-improving agents: the ability to improve task performance solely by memorizing and generalizing from past interactions. We will extend this framework to more complex, fully interleaved online learning scenarios in future work.
\end{enumerate}

\subsection{Zero-Shot Baseline Validation Experiments}
A core premise of our TTL task design is that performance improvements in the full memory setting are driven by the agent's ability to learn from historical examples, rather than prior knowledge encoded in the base LLM's pre-training. To validate this premise, we conducted zero-shot baseline evaluations, where models were tested on the TTL tasks with no access to the historical example sequence (i.e., no opportunity for test-time learning).

Table~\ref{tab:ttl_zero_shot} presents the zero-shot performance of three mainstream LLMs on our two core TTL tasks: Multi-Class Classification (MCC, Banking77) and personalized Movie Recommendation (Recom.). We also include the full memory performance of GPT-4o-mini for direct comparison, to quantify the performance degradation when test-time learning is disabled.

\begin{table}[h!]
\centering
\resizebox{0.6\textwidth}{!}{
\begin{tabular}{lccc}
\toprule
Model & MCC & Recom. & Avg.  \\
\midrule
\multicolumn{4}{l}{\textit{Zero-Shot Setting (No Access to Historical Examples)}} \\
\midrule
GPT-4o-mini & 0.6 & 6.1 & 3.4 \\
GPT-4.1-mini & 0.8 & 5.7 & 3.3 \\
Gemini-2.0-Flash & 0.0 & 5.5 & 2.8 \\
\midrule
\multicolumn{4}{l}{\textit{Full Memory Setting (With Access to Historical Examples)}} \\
\midrule
GPT-4o-mini w/ full context & 82.0 & 15.1 & 48.6 \\
\bottomrule
\end{tabular}
}
\caption{Performance on Test-Time Learning (TTL) tasks under the zero-shot setting, compared to the full memory setting. All metrics are task accuracy, with higher values indicating better performance.}
\label{tab:ttl_zero_shot}
\end{table}

The results confirm our core hypothesis: all models exhibit near-chance performance in the zero-shot setting, with average accuracy below 4\% across both tasks. In contrast, GPT-4o-mini achieves 48.6\% average accuracy when provided with the full historical example sequence, representing a 45.2 percentage point absolute improvement. This stark performance gap demonstrates that the base LLMs have no meaningful prior knowledge to solve these long-tail tasks out of the box, and all performance gains in the full memory setting are explicitly driven by the agent's ability to learn from the provided interaction history. This validates that our TTL benchmark accurately measures test-time learning capability, rather than pre-training knowledge or spurious pattern matching.

\section{Cost-Performance Analysis}
\label{app:cost_performance}

\subsection{Methodology}
To provide a realistic assessment of the practical limitations of each agent architecture, our cost calculation assumes Context Caching is enabled (a standard feature in modern APIs like OpenAI), which significantly reduces the cost for Long-Context (LC) models processing shared histories. We compare three representative architectures: Long-Context (LC) models, RAG Agents, and the agentic memory system MIRIX.

\noindent \textbf{Pricing Basis}: Costs are calculated based on OpenAI's pricing as of November 2025:
\begin{itemize}
    \item GPT-4o-mini: \$0.15 / 1M input tokens, \$0.60 / 1M output tokens
    \item GPT-4.1-mini: \$0.40 / 1M input tokens, \$1.60 / 1M output tokens
\end{itemize}
\noindent \textbf{Context Caching}: We apply cached input pricing (50\% discount for GPT-4o-mini, 75\% discount for GPT-4.1-mini) for Long-Context Agents, assuming sequential questioning on shared contexts. \\
\noindent \textbf{Settings}: RAG agents use Top-K=10. Embedding indexing costs are excluded as one-time expenses.

\subsection{Cost-Performance Results}
We report the amortized inference cost per query (in USD) and the corresponding performance metric (Accuracy/Score) across four representative datasets that vary in context length and reasoning complexity in Table~\ref{tab:cost_performance}.

\begin{table*}[h!]
\centering
\resizebox{\textwidth}{!}{
\begin{tabular}{lcccccccc}
\toprule
\multirow{2}{*}{Model/Architecture} & \multicolumn{2}{c}{MH-Doc QA} & \multicolumn{2}{c}{MCC} & \multicolumn{2}{c}{Detective QA} & \multicolumn{2}{c}{FC-SH} \\
\cmidrule(lr){2-3} \cmidrule(lr){4-5} \cmidrule(lr){6-7} \cmidrule(lr){8-9}
& Est. Cost (USD) & Performance & Est. Cost (USD) & Performance & Est. Cost (USD) & Performance & Est. Cost (USD) & Performance \\
\midrule
GPT-4o-mini & \$0.01 & 43.0 & \$0.008 & 82.0 & \$0.01 & 63.4 & \$0.01 & 45.0 \\
GPT-4.1-mini & \$0.043 & 66.0 & \$0.011 & 75.6 & \$0.013 & 56.3 & \$0.027 & 36.0 \\
RAG Agents (BM25 + 4o-mini) & <\$0.001 & 56.0 & \$0.006 & 75.4 & \$0.006 & 52.1 & <\$0.001 & 48.0 \\
MIRIX (4.1-mini) & \$0.016 & 75.0 & \$0.010 & 61.0 & \$0.011 & 62.0 & \$0.019 & 20.0 \\
\bottomrule
\end{tabular}
}
\caption{Estimated Amortized Cost vs. Performance per Query. Costs are amortized over the question set sharing the same context. Performance scores use the metrics defined in the main paper.}
\label{tab:cost_performance}
\end{table*}

\subsection{Key Insights}
\begin{enumerate}
    \item \textbf{The Efficiency-Reasoning Trade-off in RAG}: RAG agents like BM25 are cost-efficient (< \$0.001 – \$0.006 per query) but suffer in tasks requiring global reasoning (e.g., Detective QA, scoring 52.1 vs. 63.4 for LC models). This limits their utility in complex analytical scenarios.
    \item \textbf{Cost Prohibitiveness of Long-Context Scaling}: While powerful, Long-Context models face steep cost increases with stronger backbones. Upgrading from GPT-4o-mini to GPT-4.1-mini roughly quadruples the cost (from \$0.010 to \$0.043 for MH-Doc QA), making high-end long-context deployment expensive even with caching. One promising direction to mitigate such cost barriers is efficient model construction and knowledge distillation, which compress the capabilities of large models into smaller, more efficient architectures through learnable knowledge transfer (\cite{shi2024building}) and chain-based distillation (\cite{shi2026chainbaseddistillationeffectiveinitialization}). Such techniques could enable memory agents to leverage powerful backbone reasoning at significantly reduced computational cost, making long-context memory more practical for real-world deployment.
    
    \item \textbf{Agentic Memory (MIRIX) as the Optimal Middle Ground}: A critical finding is that MIRIX (with GPT-4.1-mini) achieves a lower cost (\$0.016) than the raw GPT-4.1-mini Long-Context setup (\$0.043) on memory-intensive tasks like MH-Doc QA, while delivering superior performance (75.0 vs. 66.0). This demonstrates that agentic memory mechanisms can successfully decouple performance from linear context costs, offering a scalable solution for high-performance applications.
\end{enumerate}

\section{Strict Compute-Matched Comparative Experiments}
\label{app:compute_matched}

\subsection{Experimental Setup}
To address concerns about budget effects confounding the comparison between architectures, we conducted a strict compute-matched ablation study on Banking77 (TTL) and Book Summarization (LRU) using the strongest backbone (GPT-4.1-mini). We define three budget levels:
\begin{itemize}
    \item \textbf{Low (~4K tokens)}: Constrain Long-Context (LC) models to truncated contexts; limit RAG and MIRIX to Top-K=1 chunks to match token counts.
    \item \textbf{Medium (~40K tokens)}: Moderate budget setting for all architectures; limit RAG and MIRIX to Top-K=10 chunks.
    \item \textbf{High (~100K+ tokens)}: Scale RAG and MIRIX to retrieve high Top-K chunks to match the LC full-context budget.
\end{itemize}
For Book Summarization, we evaluate on a random 30-book subset for computational efficiency. We compare total compute load via total processed tokens, rather than strictly equalizing forward passes, as forcing agentic models to a single forward pass would strip them of their core reasoning capabilities.

\begin{table*}[h!]
\centering
\resizebox{\textwidth}{!}{
\begin{tabular}{lcccc}
\toprule
Task & Budget Setting & Long-Context (GPT-4.1-mini) & RAG (BM25) & Agentic (MIRIX) \\
\midrule
\multirow{3}{*}{TTL (Banking77)} 
& Low (~4K / Top-K=1) & 74.0 & 83.0 & 52.0 \\
& Medium (~40K / Top-K=10) & 90.0 & 89.0 & 65.0 \\
& High (~104K) & 93.0 & 88.0 & 67.0 \\
\midrule
\multirow{3}{*}{LRU (Book Summarization)} 
& Low (~4K / Top-K=1) & 8.2 & 7.9 & 8.4 \\
& Medium (~40K / Top-K=10) & 16.4 & 15.8 & 18.7 \\
& High (~113K) & 39.7 & 38.0 & 38.8 \\
\bottomrule
\end{tabular}
}
\caption{Compute-matched experiment results on TTL and LRU tasks (Accuracy/Score \%).}
\label{tab:compute_matched}
\end{table*}

We summarize our key findings as follows:
\begin{itemize}
    \item \textbf{TTL: The Efficiency-Capacity Trade-off}: At the Low (~4K) budget, RAG (83.0) significantly outperforms LC models (74.0), confirming RAG's superior structural efficiency for pattern matching via precise retrieval of relevant examples, while LC models suffer heavily from truncation. As the budget increases to Medium (~40K), performance equalizes (~90.0). At the High budget, LC models scale to 93.0, whereas RAG saturates and slightly degrades (88.0) due to retrieved noise. This demonstrates that while RAG is efficient, LC models have a higher capacity ceiling when the budget permits.
    \item \textbf{LRU: The Information Threshold Effect}: For global reasoning tasks, performance exhibits a clear threshold behavior. At Low and Medium budgets, all architectures fail (<20.0 score), confirming that partial information is insufficient for full-book summarization regardless of the method. Only at the High budget (full text access) do all models achieve meaningful performance (~39.0), with RAG nearly matching LC models. This proves that for LRU tasks, success is determined by meeting the full information threshold, not by the architecture itself.
\end{itemize}

\section{Prompt Design and Overwrite Policy Ablation Experiments}
\label{app:prompt_design}

\subsection{Prompt Design Specification}
Unlike standard long-context evaluations that input raw text, we wrap all input chunks within a simulated User-Assistant dialogue to explicitly trigger the agent's memory mechanism. Each input chunk is preceded by a memorization instruction (e.g., "Please memorize the following information for future questions") to establish a clear intent for information storage. For each specific dataset, we carefully curated task instructions to ensure agents accurately comprehend the task intent and execute the required actions.

Crucially, for the Selective Forgetting competency, we introduce explicit guardrails in the prompt. We explicitly instruct agents that facts are indexed by serial numbers, and that newer facts have larger serial numbers. Agents are mandated to resolve conflicts by prioritizing the newest fact (full prompt templates are provided in the supplementary code repository).

We clarify that while prompt deltas can shift outcomes, our benchmark applies a unified and standardized prompt template across all evaluated agents (Long-context, RAG, and Agentic). This ensures that the performance gaps observed (e.g., the failure of RAG agents in multi-hop Selective Forgetting) are attributed to the limitations of their memory mechanisms rather than prompt inconsistency.

\subsection{Overwrite Policy Ablation Experiments}
To rigorously test whether explicit instructions can mitigate the forgetting/overwriting issue, we conducted additional ablation studies using explicit overwrite prompts on the GPT-4.1-mini baseline. We test two policy settings:
\begin{itemize}
    \item \textbf{Policy A (Always Prefer Later)}: "Crucial Rule: Treat the facts as a chronological update stream. If there is ANY conflict between facts, you must ALWAYS overwrite the earlier fact with the one having the larger serial number."
    \item \textbf{Policy B (Conservative/Explicit Negation)}: "Crucial Rule: Be conservative with updates. ONLY discard or overwrite an earlier fact if the fact with the larger serial number explicitly negates it or explicitly states the previous information is incorrect."
\end{itemize}

\subsubsection{Ablation Results}
The results of the overwrite policy ablation are presented in Table~\ref{tab:overwrite_policy}.
\begin{table}[h!]
\centering
\begin{tabular}{lccc}
\toprule
Model/Setting & FC-SH & FC-MH & Avg. \\
\midrule
GPT-4.1-mini (Baseline) & 36.0 & 5.0 & 20.5 \\
GPT-4.1-mini (Policy A) & 40.0 & 4.0 & 22.0 (+1.5) \\
GPT-4.1-mini (Policy B) & 28.0 & 4.0 & 16.0 \\
\bottomrule
\end{tabular}
\caption{Overwrite policy ablation results on Selective Forgetting tasks (Accuracy \%).}
\label{tab:overwrite_policy}
\end{table}

\subsubsection{Key Insights}
\begin{enumerate}
    \item \textbf{Limited Generalization with Aggressive Updates}: While Policy A slightly improves performance on single-hop tasks (FC-SH score increases from 36.0 to 40.0), it fails to generalize to complex multi-hop reasoning (FC-MH drops to 4.0). This suggests that while prompting helps with simple fact retrieval, it cannot effectively propagate updates through multi-step reasoning chains.
    \item \textbf{Performance Degradation with Conservative Constraints}: Policy B significantly degrades average performance (-4.5 points). The complex conditional instruction (checking for explicit negation) increases the model's cognitive load and induces overly cautious behavior, preventing the retrieval of valid updates.
\end{enumerate}
These findings serve as a sanity check that validates our core motivation: Selective Forgetting cannot be solved by prompt engineering alone, and requires dedicated memory mechanism design.

\section{Supplementary Notes for LLM-as-a-Judge and Input Format}
\label{app:additional_notes}

\subsection{Validity of LLM-as-a-Judge Evaluation}
We acknowledge that model-based evaluation may not be optimal for all task contexts. However, we adopted LLM-as-a-judge scoring on the LongMemEval and $\infty$Bench-Sum datasets to remain consistent with prior work, and we validate its appropriateness as follows:
\begin{itemize}
    \item For LongMemEval, the questions admit clear, objective ground-truth answers with minimal subjectivity. ~\citet{longmemeval} report that a prompt-engineered GPT-4o judge achieves 98.0\% agreement with human annotations, demonstrating very high stability.
    \item ~\citet{HELMET} validate that GPT-4o judgments mostly align with human evaluations for long-context summarization tasks.
\end{itemize}
These findings confirm that our LLM-as-a-judge setup reliably reflects human evaluation for the tasks in our benchmark.

\subsection{Rationale for Chunked Input Format}
We acknowledge that real-world personal assistants may receive streaming input (e.g., continuous user interactions or real-time data streams). However, in practice, it is necessary to quantize real-world inputs before feeding them into a language model (e.g., accumulating continuous input into discrete chunks for inference). Thus, inputting chunks into the agent is a natural and realistic strategy to handle streaming input in real-world deployments.

Additionally, the chunked input format simulates the incremental, multi-turn nature of real-world user-agent interactions, where information arrives sequentially over time rather than in a single full document. This setup is critical for evaluating memory agents, which are designed to process information incrementally, rather than the static full-document input used in standard long-context benchmarks.